\documentclass[letterpaper]{article} 
\usepackage{aaai2026}  
\usepackage{times}  
\usepackage{helvet}  
\usepackage{courier}  
\usepackage[hyphens]{url}  
\usepackage{graphicx} 
\usepackage{natbib}  
\usepackage{caption} 
\usepackage{amsmath, amsfonts, amsthm}
\usepackage{subfiles}
\usepackage{bbold}
\usepackage{multirow, booktabs, graphicx}
\usepackage{xcolor, colortbl}
\usepackage{subfig}
\usepackage{makecell}

\usepackage{cuted}

\definecolor{1st}{RGB}{220,  38, 127}
\definecolor{2nd}{RGB}{0  , 175,  95}
\definecolor{color1}{RGB}{238, 252, 241}    
\definecolor{color2}{RGB}{253, 228, 233}    
\definecolor{color3}{RGB}{255, 245, 225}    
\definecolor{color4}{RGB}{232, 244, 253}    
\definecolor{gray}{RGB}{229, 229, 229}

\usepackage{algorithm}
\usepackage{algorithmic}

\usepackage{newfloat}
\usepackage{listings}
\DeclareCaptionStyle{ruled}{labelfont=normalfont,labelsep=colon,strut=off} 
\floatstyle{ruled}
\newfloat{listing}{tb}{lst}{}
\floatname{listing}{Listing}
\title{From Entanglement to Alignment: Representation Space Decomposition for Unsupervised Time Series Domain Adaptation}
\author {
    Rongyao Cai\textsuperscript{\rm 1},
    Ming Jin\textsuperscript{\rm 2},
    Qingsong Wen\textsuperscript{\rm 3},
    Kexin Zhang\textsuperscript{\rm 1}\thanks{Correspondence to: Kexin Zhang (zhangkexin@zju.edu.cn).}
}
\affiliations {
    \textsuperscript{\rm 1}Insititute of Cyber-Systems and Control, Zhejiang University, Hangzhou, China\\
    \textsuperscript{\rm 2}School of Information and Communication Technology, Griffith University, Brisbane, Australia\\
    \textsuperscript{\rm 3}Squirrel Ai Learning, Bellevue, USA\\
    \{rycai, zhangkexin\}@zju.edu.cn, ming.jin@griffith.edu.au, qingsongedu@gmail.com
}

\begin{document}

\maketitle

\begin{abstract}
    Domain shift poses a fundamental challenge in time series analysis, where models trained on source domain often fail dramatically when applied in target domain with different yet similar distributions. While current unsupervised domain adaptation (UDA) methods attempt to align cross-domain feature distributions, they typically treat features as indivisible entities, ignoring their intrinsic compositions that govern domain adaptation.
    We introduce \textbf{DARSD}, a novel UDA framework with \textbf{\textit{theoretical explainability}} that explicitly realizes UDA tasks from the perspective of \textbf{\textit{representation space decomposition}}. Our core insight is that effective domain adaptation requires not just alignment, but principled disentanglement of transferable knowledge from mixed representations. DARSD consists of three synergistic components: \textbf{(I)} An adversarial learnable common invariant basis that projects original features into a domain-invariant subspace while preserving semantic content; \textbf{(II)} A prototypical pseudo-labeling mechanism that dynamically separates target features based on confidence, hindering error accumulation; \textbf{(III)} A hybrid contrastive optimization strategy that simultaneously enforces feature clustering and consistency while mitigating emerging distribution gaps.
    Comprehensive experiments conducted on four benchmarks (WISDM, HAR, HHAR, and MFD) demonstrate DARSD's superiority against 12 UDA algorithms, achieving optimal performance in 35 out of 53 scenarios and  ranking first across all benchmarks. 
\end{abstract}

\section{Introduction}
In the context of rapid development of time series analysis, domain adaptation has emerged as a critical bottleneck of practical algorithm deployment across diverse environments~\cite{10679601, FANG2024106230}.
Consider a human activity recognition system trained on smartphone accelerometer data from young adults in a lab, its accuracy can drop sharply when used on smartwatches worn by elderly users in daily life~\cite{10919169}. This dramatic performance degradation stems from a fundamental issue: while the underlying patterns of human activities stay the same, like walking still exhibits periodic acceleration patterns, the data distributions shift significantly due to fluctuations in objects, environments, or perception~\cite{9288927}. Such domain shift scenarios are widespread across real-world applications ranging from industrial equipment monitoring~\cite{10600455, 10879141} to healthcare sensing~\cite{YE2024110811}.

Due to a lack of expert knowledge or high cost~\cite{ozyurt2023contrastive, 10443248}, the ground truth of target domain is not available during training in most scenarios. Unsupervised domain adaptation (UDA) has gained increasing attention, which trains a robust model using annotated source data with unlabeled target data, aiming to achieve good performance on the latter~\cite{10679601, Westfechtel_2024_WACV}. Notably, source and target domain exhibit similar yet distinct distributions.

\begin{figure}
    \centering
    \includegraphics[width=1.0\linewidth]{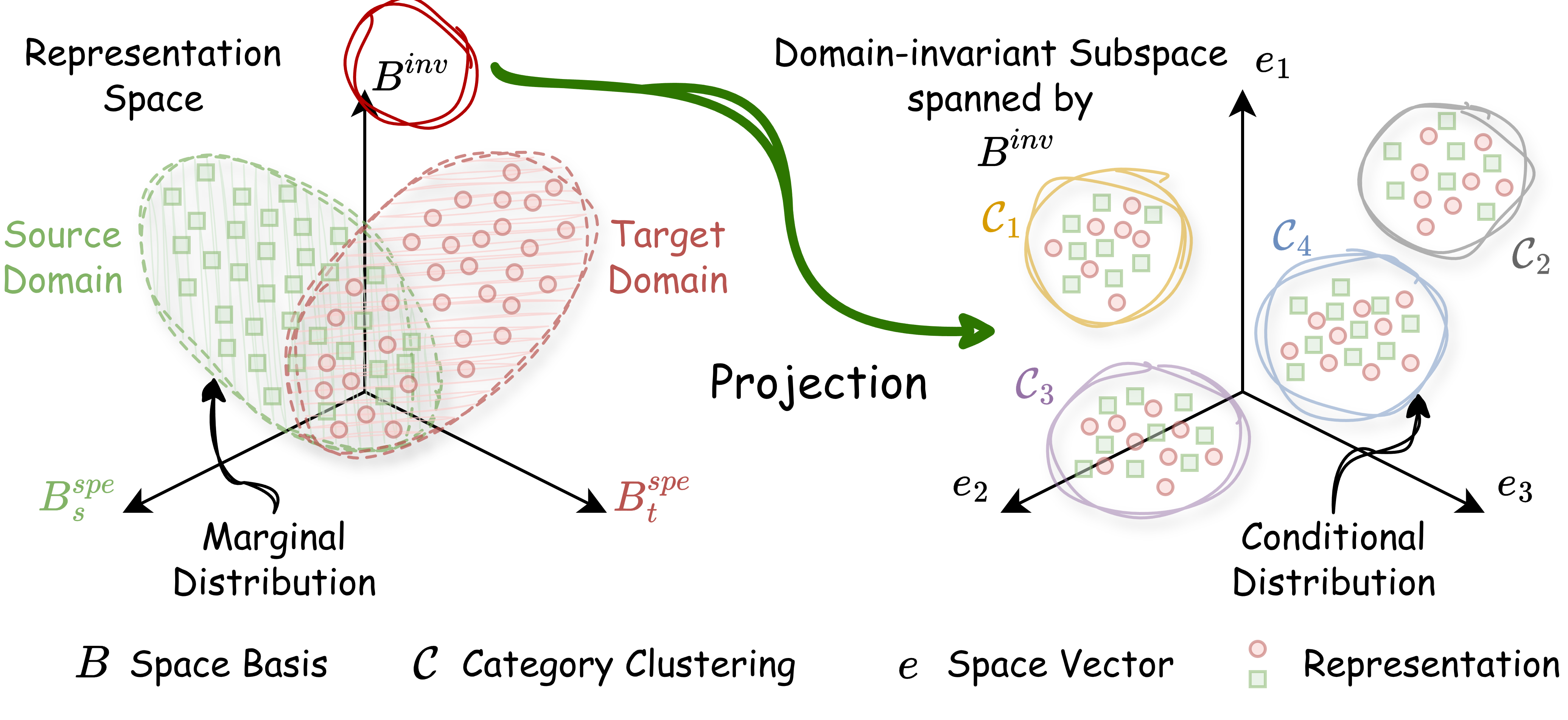}
    \caption{Visual description of representation space decomposition. We disentangle the basis of representation space into invariant one $B^{\textit{inv}}_{}$ and specific ones related to domains $B^{\textit{spe}}_s$ and $B^{\textit{spe}}_t$, aiming to achieve discriminative clustering in domain-invariant subspace spanned by $B^{\textit{inv}}$.}
    \label{fig:abs}
\end{figure}

Existing UDA approaches attempt to address domain shift through feature distribution alignment, they can be classified into three main categories.
\textbf{\textit{Adversarial training methods}}~\cite{10688076, XIE2025109668} learn domain-confusing representations by fooling discriminators~\cite{9950721}. But this adversarial game often lacks supervision linked to downstream tasks, eliminating meaningful patterns alongside domain-specific artifacts.
\textbf{\textit{Metric learning approaches}}~\cite{9085896, 10979996} minimize statistical distances between source and target distributions~\cite{9950721}, yet they impose an unrealistic assumption that entire features need be aligned, despite the fact that only part of components carry transferable knowledge. 
\textbf{\textit{Self-supervised methods}}~\cite{XU2025125452, 10.5555/3618408.3618926} depend on contrastive learning with pretext tasks, such as pseudo-labeling~\cite{chen2021transferrable}. However, they run into a problem that label noise tends to build up progressively over the course of training. 
The core flaw underlying above approaches is that they treat features as \textbf{monolithic and indivisible entities}, implicitly hoping for invariance through distribution matching, overlooking the fact that effective domain adaptation demands \textit{a principled distinction between what ought to transfer and what ought not to}. Though several methods have been proposed based on the insight of feature decomposition, they rely on the implicit network structure~\cite{10.24963/ijcai.2024/424, Qu_2024_CVPR} or causal hidden variables~\cite{10.5555/3692070.3693414}.

Our work is motivated by a key insight from representation learning: \textit{features possess internal structure}, as depicted in Fig.~\ref{fig:abs}. Domain shift primarily impacts certain dimensions of representations while leaving others unaltered. For instance, in accelerometer data, the temporal dynamics of walking remain largely consistent across devices (\textit{domain-invariant}), whereas sensor noise characteristics and sampling artifacts vary significantly (\textit{domain-specific}). This observation suggests a conceptual shift, \textit{instead of aligning entire feature distributions, we should explicitly decouple representations into orthogonal subspaces and selectively transfer only the invariant components}. Such decomposition not only preserves semantic information but also provides interpretable insights into which patterns actually transfer across domains.

Based on this perspective, we introduce \textbf{DARSD (\underline{D}omain \underline{A}daptation via \underline{R}epresentation \underline{S}pace \underline{D}ecomposition)}, a novel UDA framework that explicitly disentangles domain-invariant from domain-specific feature components. 
To achieve this goal, DARSD employs three synergistic innovations: 
\textbf{(I) \textit{Adversarial Learnable Common Invariant Basis (Adv-LCIB)}}: a learnable orthogonal transformation that projects source and target features into a shared domain-invariant subspace while adversarially remaining the information completeness; 
\textbf{(II) \textit{Prototypical Pseudo-label Generation with Confidence Evaluation (PPGCE)}}: a dynamic mechanism that assigns reliable pseudo-labels to target features, partitions them by confidence, and thereby prevents the accumulation of label noise; 
\textbf{(III) \textit{Hybrid Contrastive Optimization}}: a carefully designed collaborative optimization architecture that balances feature clustering for discrimination with consistency enforcement for robustness.
Briefly put, our key contributions are summarized as follows:
\begin{itemize}
    \item \textbf{\textit{Innovative Paradigm}:} To the best of our knowledge, \textbf{DARSD} is the first explicit representation space decomposition framework for time series UDA tasks.
    \item \textbf{\textit{Interpretable Theory}:} We propose a novel Adv-LCIB mechanism with theoretical foundations to conduct feature projection and domain-invariant pattern extraction.
    \item \textbf{\textit{Robust Optimization}:} We introduce a hybrid contrastive optimization strategy integrated with the PPGCE module to implement differentiated treatment of cross-domain features, creating homogeneous yet discriminative category-wise feature distributions.
    \item \textbf{\textit{Experimental Evaluation}:} Comprehensive experiments on four benchmarks against 12 UDA algorithms demonstrate the superior performance of our DARSD framework, achieving 35 out of 53 SOTA results.
\end{itemize}

\begin{figure*}
    \centering
    \includegraphics[width=0.9\linewidth]{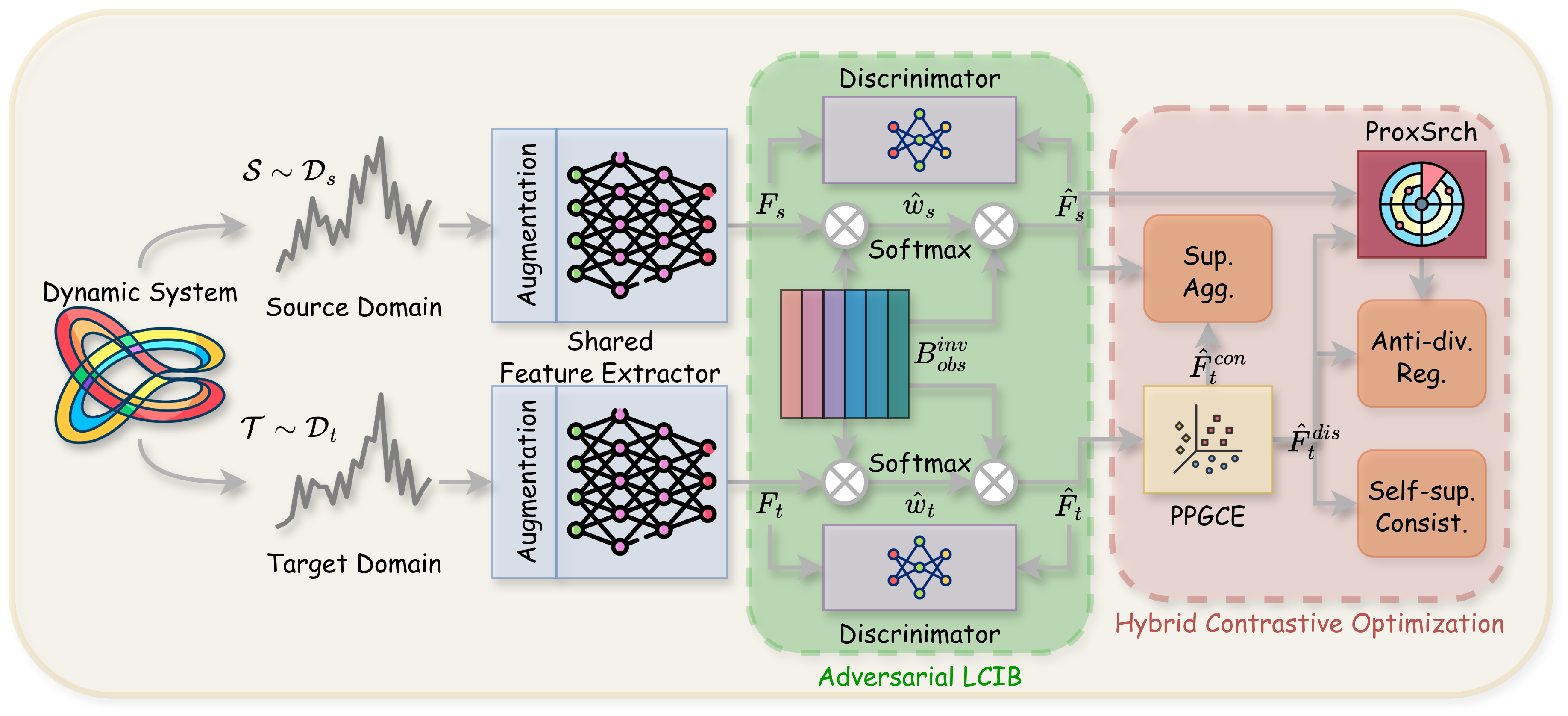}
    \caption{The DARSD framework for time series domain adaptation. Features are first projected into a domain-invariant subspace (Adv-LCIB), then target features receive pseudo-labels with confidence evaluation (PPGCE), and finally all features undergo hybrid contrastive optimization for category-wise aggregation across domains. In hybrid contrastive optimization block, (1) \textit{Sup. Agg.}, (2) \textit{Self-sup. Consist.}, (3) \textit{Anti-div. Reg.}, and (4) \textit{ProxSrch} are the abbreviations of (1) \textit{Supervised Aggregation}, (2) \textit{Self-supervised Consistence}, (3) \textit{Anti-divergence Regularization}, and (4) \textit{Proximity Searcher}, respectively.}
    \label{fig:arch}
\end{figure*}

\section{Problem Definition}
\label{sec:problem definition}

UDA for time series essentially boils down to a classification task involving samples from divergent distributions. 
Given an annotated source dataset $\mathcal{S}$ and an unannotated target dataset $\mathcal{T}$, which follow similar yet different distributions, the objective of UDA is to predict the labels of $\mathcal{T}$. Specifically, $\mathcal{S}$ is denoted as  $\{ (x^s_{i}, y^s_{i}) \}^{n_s}_{i=1} \sim \mathcal{D}_s$, comprising $n_{s}$ samples drawn from distribution $\mathcal{D}_{s}$. Here, $x^s_{i} \in \mathbb{R}^{T \times D}$ represents a source sample capturing $D$ sensors over $T$ time points and each label $y^s_{i}$ belongs to the label set $\mathcal{C}$ with $n_c$ categories. $\mathcal{T}$ is defined as $\{ x^t_{i} \}^{n_t}_{i=1} \sim \mathcal{D}_t$ with $n_{t}$ samples following distribution $\mathcal{D}_{t}$, where $x^t_{i} \in \mathbb{R}^{T \times D}$ denotes a target sample with unknown label $y^t_{i} \in \mathcal{C}$.

Our objective is to train a robust feature extractor $\textit{FE}(\cdot)$ by leveraging annotated $\mathcal{S}$ and unlabeled $\mathcal{T}$. This extractor will derive a domain-invariant feature set $F=\{ F_s \cup F_t \}^{n_s+n_t}$ from $\mathcal{S}$ and $\mathcal{T}$. Subsequently, we fine-tune a classifier $\textit{CLF}(\cdot)$ using $\{ (f^s_{i}, y^s_{i}) \}_{i=1}^{n_s}$ to predict the label set $Y_t = \{ y^t_{i} \}_{i=1}^{n_t}$ for $\mathcal{T}$ based on the feature set $F_t = \{ f^t_{i} \}_{i=1}^{n_t}$.

\section{Methodology}
\label{sec:method}
The detailed pre-training procedure of our DARSD framework is illustrated in Fig.~\ref{fig:arch}. DARSD comprises a shared feature extractor $\textit{FE}(\cdot)$, an Adv-LCIB module, and a hybrid contrastive optimization mechanism integrated with the PPGCE module.

\subsection{Adversarial Learnable Common Invariant Basis}
The fundamental insight guiding our approach is that features are compositional that they can be decomposed into orthogonal components serving distinct functional roles. In time series data, specific feature dimensions capture semantic patterns consistent across domains ($e.g.$, temporal correlations, periodic structures, trend characteristics), whereas others encode domain-varying environmental artifacts ($e.g.$, sampling noise, device-specific biases, environmental interference). Traditional UDA methods seek to align entire feature distributions, which inadvertently mixes these heterogeneous components, compromising both semantic preservation and domain invariance.

Our key innovation lies in explicitly achieving decomposition via Adv-LCIB module. Rather than relying on implicit alignment to preserve semantic content, we construct an observed learnable orthogonal basis $B^{\textit{inv}}_{\textit{obs}} \in \mathbb{R}^{d \times m}$ that spans a domain-invariant subspace. This basis acts as a semantic filter, projecting features into a subspace that captures transferable patterns while excluding domain-specific artifacts.

\textbf{Feature Decomposition Principle}: Any feature $f \in \mathbb{R}^d$ regardless of domains can be conceptually decomposed as: 
\begin{equation}
    f = \textit{FE}(x) = f^{\textit{inv}} + f^{\textit{spe}}
\end{equation}
where $f^{\textit{inv}}$ contains domain-invariant semantic patterns and $f^{\textit{spe}}$ contains domain-specific artifacts. Our goal is to extract $f^{\textit{inv}}$ without explicit prior knowledge of this decomposition.

\textbf{Learnable Basis Construction}: We introduce a learnable orthogonal matrix $B^{\textit{inv}}_{\textit{obs}} \in \mathbb{R}^{d \times m}$, spanning the domain-invariant subspace. For any given feature $f$, its domain-invariant representation is calculated as follows:
\begin{gather}
    \begin{aligned}
        w^{\textit{inv}} &= (B^{\textit{inv}}_{\textit{obs}})^{\top} f \\
          &= (B^{\textit{inv}}_{\textit{obs}})^{\top} (f^{\textit{inv}} + f^{\textit{spe}}) \\
          &= (B^{\textit{inv}}_{\textit{obs}})^{\top} f^{\textit{inv}} = (B^{\textit{inv}}_{\textit{obs}})^{\top} B^{\textit{inv}} w^{\textit{inv}} 
        \label{eq:inv_coordinate}
    \end{aligned} \\
    \hat{f} =  B^{\textit{inv}}_{\textit{obs}} w^{\textit{inv}}
\end{gather}
where $w^{\textit{inv}}$ denotes the coordinate of $f$ in domain-invariant subspace spanned by $B^{\textit{inv}}$. $B^{\textit{inv}}_{\textit{obs}}$ is the observed approximation of authentic invariant space basis $B^{\textit{inv}}$. The \textbf{\textit{proof of} Eq.~\eqref{eq:inv_coordinate}} is depicted in \textbf{Appendix B}.

During training process, $B^{\textit{inv}}_{\textit{obs}}$ gradually approaches $B^{\textit{inv}}$ but may inadvertently incorporate some domain-specific artifacts. Given that invariant patterns dominate $B^{\textit{inv}}_{\textit{obs}}$ while domain-specific artifacts account for a smaller proportion, DARSD employs the softmax regularization to amplify the weight differences between them. This operation suppresses domain-specific noise, thereby ensuring the efficiency of invariant information in the reconstructed features $\hat{f}$.
\begin{gather}
    \hat{w} = \text{Softmax}(w^{\textit{inv}}) = \text{Softmax}( (B^{\textit{inv}}_{\textit{obs}})^{\top} \textit{FE}(x)) \\
    \hat{f} = B^{\textit{inv}}_{\textit{obs}} \hat{w} = B^{\textit{inv}}_{\textit{obs}} \text{Softmax}( (B^{\textit{inv}}_{\textit{obs}})^{\top} \textit{FE}(x)) 
\end{gather}

Overall, LCIB module consists of the following stages: 
\textbf{(1) Projection}: $B^{\textit{inv}}_{\textit{obs}}$ projects feature $f$ onto the invariant subspace to obtain the coordinate $w^{\textit{inv}}$. 
\textbf{(2) Softmax Regularization}: Ensures only the most relevant invariant components are retained, naturally suppressing domain-specific noise and yielding $\hat{w}$. 
\textbf{(3) Reconstruction}: $B^{\textit{inv}}_{\textit{obs}}$ reconstructs the domain-invariant representation $\hat{f}$ using coordinate $\hat{w}$.

\textbf{Adversarial Solution}: LCIB is agnostic to the quality of $B^{\textit{inv}}_{\textit{obs}}$, whether it truly captures domain-invariant patterns or merely memorizes input features. A naive, unconstrained basis might inadvertently retain unintended domain-specific noise while missing out on expected domain-invariant patterns, defeating the purpose of decomposition. 

To tackle this issue, we analyze the proportion of $f^{\textit{inv}}$ and $f^{\textit{spe}}$ within $f$. $f^{\textit{spe}}$ primarily captures distribution shifts induced by dynamic sampling environments, whereas $f^{\textit{inv}}$ encodes essential characteristics that remain consistent across domains. Given their critical role in domain adaptation, we hypothesize that $f^{\textit{inv}}$ should govern features $f$. To this end, DARSD incorporates a discriminator $D(\cdot): \mathbb{R}^d \rightarrow [0,1]$ tasked with distinguishing between original features $f$ and corresponding reconstructed counterparts $\hat{f}$ with the goal of keeping $f^{\textit{inv}}$ as much as possible. The discriminator is trained to maximize:
\begin{equation}
    \mathcal{L}^{\textit{adv}} = \mathbb{E}_{\bar{f}_i \in \{ F \cup \hat{F}\}} \Bigg[ z_i \log(D(\bar{f}_i)) + (1 - z_i) \log(1 - D(\bar{f}_i)) \Bigg]
\end{equation}
where $z_i$ is assigned label for $\bar{f}_i \in \{ F \cup \hat{F}\}$.

\subsection{Prototypical Pseudo-label Generation with Confidence Evaluation}
The domain-invariant feature sets $\hat{F}_s$ and $\hat{F}_t$ derived from Adv-LCIB create an opportunity for effective cross-domain knowledge transfer. However, a critical asymmetry persists: source features are accompanied by ground-truth labels, enabling supervised learning and category-wise clustering, while target features lack supervisory signals and thus cannot directly participate in discriminative optimization.

To fully leverage target features, assigning reliable pseudo-labels to $\hat{F}_t$ is a viable solution, allowing them to join source features in supervised learning. This enables category-wise aggregation, where features from the same semantic class are clustered together regardless of domains, while those from different classes are separated. Even so, pseudo-labels vary in trustworthiness, that unreliable ones can introduce noise, degrade the quality of clustering, and disrupt decision boundaries. We need a mechanism to distinguish reliable from unreliable pseudo-labels and treat them differently during optimization.

Rather than relying on classifier predictions that may be biased toward source domain characteristics, we generate pseudo-labels through prototypical similarity in the domain-invariant subspace. This approach capitalizes on the geometric structure of reconstructed invariant representations.

\textbf{Momentum-based Prototype Construction}: For each class $c \in \mathcal{C}$, we maintain a prototype $p_c$ that serves as the centroid of class $c$ within the domain-invariant subspace:
\begin{equation}
    p_c^{t+1} = \mu p_c^{t} + (1-\mu)\frac{1}{| \mathcal{S}_c |} \sum_{\hat{f}^s_i \in \mathcal{S}_c} \hat{f}^s_i  
\end{equation}
where $\mathcal{S}_c = \{ \hat{f}_i | y^s_i = c, \hat{f}_i \in \hat{F}_s \}$ contains source features of class $c$, and $\mu$ controls the momentum factor.

\textbf{Pseudo-label Assignment}: For each target feature $\hat{f}_j^t$, we assign pseudo-label based on maximum cosine similarity:
\begin{align}
    y_{i}^{\textit{psd}} &= \arg \max_{c} \cos(\hat{f}^t_i, p_c)
    \\
    \sigma_{i} &= \max_{c} \cos(\hat{f}^t_i, p_c)
\end{align}
where $\sigma_i$ serves as the confidence score for the assigned pseudo-label $y_i^{\textit{psd}}$.

\textbf{Adaptive Thresholding}: Instead of using a fixed confidence threshold, we employ a curriculum learning strategy with increasing confidence ratio $\eta(t)$:
\begin{equation}
    \eta(t) = \eta(0) + \frac{t}{T_{\textit{total}}} \cdot (\eta_{\textit{max}} - \eta(0))
\end{equation}

This progressive inclusion strategy strikes a balance between exploration (utilizing more target data) and exploitation (maintaining pseudo-label quality).

\textbf{Confidence-based Partitioning}: At each training step, target features are separated into two parts:
\begin{itemize}
    \item Confident subset: $\hat{F}_t^{\textit{con}} = \{ (\hat{f}_i^t, y_i^{\textit{psd}}) | \sigma_i \geq \eta(t) \}$;
    \item Distrusted subset: $\hat{F}_t^{\textit{dis}} = \hat{F}_t \setminus \hat{F}^{\textit{con}}_t$.
\end{itemize}

The dynamic partitioning serves two purposes: \textbf{(I)} Confident features directly participate in supervised aggregation with source data; \textbf{(II)} distrusted features undergo self-supervised consistency training, gradually improving their reliability to qualify for future supervised aggregation.

\subsection{Hybrid Contrastive Optimization}
At this stage, we perform operations on three distinct types of features: labeled source features $\hat{F}_s$, confident target features $\hat{F}_t^{\textit{con}}$ with reliable pseudo-labels, and distrusted target features $\hat{F}_t^{\textit{dis}}$. The ultimate goal is to achieve effective category-wise aggregation, where same-category features are gathered and those from different classes are separated.

A straightforward approach would be to apply standard supervised contrastive learning to $\hat{F}_s$ and $\hat{F}_t^{\textit{con}}$ while discarding distrusted $\hat{F}_t^{\textit{dis}}$. Nevertheless, this strategy suffers from three fundamental limitations that compromise adaptation effectiveness.
\textbf{(1) Underutilization of Target Data}: Discarding $\hat{F}_{t}^{\textit{dis}}$ represents a significant waste of target domain information, which could contribute to better representation learning;
\textbf{(2) Emerging Distribution Divergence}: Focusing solely on confident target features creates a emerging distribution gap between the supervised subset $\{\hat{F}_s \cup \hat{F}_t^{\textit{con}} \}$ and the excluded $\hat{F}_t^{\textit{dis}}$, potentially undermining the domain-invariant property established by the Adv-LCIB module;
\textbf{(3) Conflicting Optimization Objectives}: Without proper coordination, contradictory optimization pressures may arise between the discriminative clustering objective and the robust consistency objective.

To address above issues, we propose a hybrid contrastive optimization strategy that simultaneously leverages all three feature types through complementary loss functions. 
Instead of confining all features into a single optimization framework, we design specialized objectives that exploit the unique characteristics of each feature type while ensuring overall coherence. The supervised aggregation loss $\mathcal{L}_{\textit{sup}}$ as dominant constraint handles the reliable features to form robust discriminative clusters; the self-supervised consistency loss $\mathcal{L}_{\textit{self}}$ gradually enhances the confidence of distrusted target features; and the anti-divergence regularization $\mathcal{L}_{\textit{anti}}$ bridges the distribution gap between these two pathways.

\textbf{Supervised Aggregation $\mathcal{L}_{\textit{sup}}$}: 
To enable direct semantic category-wise aggregation between source features and confident target features, DARSD treats $\hat{F}_s$ and $\hat{F}_t^{\textit{con}}$ equally and conducts supervised contrastive learning on them using reliable labels. Specifically, for any feature $\hat{f}_i \in \{ \hat{F}_s \cup \hat{F}_t^{\textit{con}} \}$, we define the positive set $\mathcal{P}(i)$ as features sharing the same class label and its negative set $\mathcal{N}(i)$ as features from different classes. The supervised loss is formulated as:
\begin{equation}
\begin{aligned}
    &\mathcal{L}_{\textit{sup}} = \\
    &\mathbb{E}_{\hat{f}_i \in \{ \hat{F}_s \cup \hat{F}^{\textit{con}}_t \}}
    \Bigg[ \mathbb{E}_{\hat{f}_p \in \mathcal{P}(i)}
    - \log 
    \frac{\exp (\cos (\hat{f}_i, \hat{f}_p) / \tau)}
         {\sum_{\hat{f}_n \in \mathcal{N}(i)} \exp (\cos (\hat{f}_i, \hat{f}_n) / \tau)}
    \Bigg]
\end{aligned}
\end{equation}
where $\tau$ is the temperature parameter. 

\textbf{Self-supervised Consistency $\mathcal{L}_{\textit{self}}$}: 
For the distrusted target features, we employ self-supervised consistency learning to enhance their confidence rather than immediately introducing noise into the supervised aggregation, making them candidates for promotion to the confident set in future iterations. Each distrusted feature $\hat{f}_i^{\textit{dis}}$ is paired with its augmented view $\hat{f}_i^{\textit{dis}+}$ to form positive pairs, while other distrusted features serve as negatives. This consistency objective maintains representational stability for distrusted features while preventing overfitting to potentially noisy pseudo-labels.
\begin{equation}
    \mathcal{L}_{\textit{self}} =
    \mathbb{E}_{\hat{f}^\textit{dis}_{i} \in \hat{F}^{\textit{dis}}_{t}}
    \Bigg[
        -\log 
        \frac{\exp (\cos (\hat{f}_{i}^{\textit{dis}}, \hat{f}_{i}^{\textit{dis}+}) / \tau)}
             {\sum_{j \neq i} \exp (\cos (\hat{f}_{i}^{\textit{dis}}, \hat{f}_{j}^{\textit{dis}}) / \tau)}
    \Bigg]
\end{equation}

\textbf{Anti-divergence regularization $\mathcal{L}_{\textit{anti}}$}: 
A critical issue arises with the emerging distribution divergence caused by supervised aggregation and self-supervised consistency. Without proper coordination, these two optimization pathways may yield distinct feature distributions, violating the domain-invariant assumption and disrupting pseudo-label generation. Specifically, supervised aggregation produces cluster-wise distribution, while self-supervised consistency tends to generate uniform one~\cite{10446381}. 
To address this, we introduce an anti-divergence regularization that links distrusted target features to their semantically closest source counterparts. For each $\hat{f}_i^{\textit{dis}}$, we identify its nearest semantic neighbor in the source domain by proximity searcher operator $\textit{PS}(\cdot)$: 
\begin{equation}
    \textit{PS}(\hat{f}^{\textit{dis}}_i, \hat{F}_s) = \arg \max_{{\hat{f}^s_j \in \hat{F}_s}} \cos(\hat{f}^{\textit{dis}}_i, \hat{f}^s_j)
\end{equation}

The anti-divergence regularization loss then pulls each distrusted feature toward its identified source anchor, bridging mechanism prevents distribution heterogeneity of distrusted $\hat{F}_t^{\textit{dis}}$ and constrain distributional coherence across all feature types.
\begin{equation}
    \mathcal{L}_{\textit{anti}} = 
    \mathbb{E}_{\hat{f}^{\textit{dis}}_{i} \in \hat{F}^{\textit{dis}}_{t}}
    \Bigg[
        -\log 
        \frac{\exp (\cos (\hat{f}_{i}^{\textit{dis}}, \textit{PS}(\hat{f}_{i}^{\textit{dis}}, \hat{F}_s)) / \tau)}
             {\sum_{\hat{f}^s_{j} \in \hat{F}_s} \exp (\cos (\hat{f}_{i}^{\textit{dis}}, \hat{f}^s_{j}) / \tau)}
    \Bigg]
\end{equation}

Hybrid contrastive optimization creates a virtuous cycle where better clustering leads to more reliable prototypes, which in turn generate higher-quality pseudo-labels and expand the confident target pool. Meanwhile, improved consistency and distribution bridge provides more stable target features, promoting the conversion of distrusted target features into confident ones and reducing evolving distribution divergence. The final objective combines all these pathways with the adversarial loss from Adv-LCIB module:
\begin{equation}
    \label{eq:hybrid_loss}
    \mathcal{L}_{\textit{total}} = \mathcal{L}_{\textit{sup}} + \mathcal{L}_{\textit{self}} + \lambda_1 \mathcal{L}_{\textit{anti}} + \lambda_2 \mathcal{L}_{\textit{adv}}
\end{equation}
where $\lambda_1$ and $\lambda_2$ are balancing hyperparameters.

\section{Experimental Evaluation}
\label{sec:experiments}

\subsection{Benchmarks and Evaluation Metrics}

\begin{table}[]
\centering
\small
\begin{tabular}{c|ccc|ccc}
\bottomrule

\toprule
Datasets  & $D$  & $T$    & $n_c$ & Subject & Train  & Test \\ \midrule
WISDM     & 3    & 128    & 6   & 30      & 1350   & 720  \\
HAR       & 9    & 128    & 6   & 30      & 2300   & 990  \\
HHAR      & 3    & 128    & 6   & 9       & 12716  & 5218 \\
MFD       & 1    & 5120   & 3   & 4       & 7312   & 3604 \\
\bottomrule

\toprule
\end{tabular}
\caption{Characteristics of benchmarks.}
\label{tab:dataset}
\end{table}

To demonstrate the effectiveness of our DARSD, we conduct comprehensive evaluations on four widely-used real-world benchmarks: \textbf{WISDM}~\cite{10.1145/1964897.1964918}, \textbf{HAR}~\cite{Anguita2013APD}, \textbf{HHAR}~\cite{stisen2015smart}, and \textbf{MFD}~\cite{lessmeier2016condition}. These datasets represent diverse application domains. Detailed characteristics of each dataset are summarized in Table~\ref{tab:dataset} and \textbf{Appendix C}.

We employ the \textbf{Macro-F1} score as primary evaluation metric for UDA tasks. This metric offers a comprehensive assessment of model performance by calculating the unweighted average of F1 scores across all classes, thus ensuring that both frequent and rare classes receive equal consideration in the evaluation process.

\subsection{Baselines}
We compare proposed DARSD framework against 12 state-of-the-art baselines. Based on their primary technical contributions, we categorize them into three distinct groups: \\
\textbf{(1) \textit{Adversarial approaches}}: These methods leverage adversarial training mechanisms to align domain distributions, including VRADA~\cite{purushotham2017variational}, CDAN~\cite{10.5555/3326943.3327094}, CoDATS~\cite{wilson2020multi}, and CADT~\cite{10.24963/ijcai.2024/424}; \\
\textbf{(2) \textit{Metric learning approaches}}: These techniques focus on learning domain-invariant representations through various distance metrics and alignment strategies, encompassing DDC~\cite{tzeng2014deep}, D-CORAL~\cite{10.1007/978-3-319-49409-8_35}, CAN~\cite{kang2019contrastive}, HoMM~\cite{chen2020homm}, MMDA~\cite{Rahman2020}, AdvSKM~\cite{ijcai2021p378}, and DSAN~\cite{9085896}, and DARSD (ours); \\
\textbf{(3) \textit{Self-supervised approaches}}: This category utilizes self-supervised learning paradigms for domain adaptation, represented by CLUDA~\cite{ozyurt2023contrastive}.

\subsection{Implementation Details}
To ensure fair comparison across all methods, we adopt a consistent network architecture where the feature extractor is a Temporal Convolutional Network (TCN) with 4 layers, 128 hidden dimensions, 0.2 dropout ratio, and produces feature representations of length $d=128$. The classification head consists of a two-layer MLP with 128 hidden dimensions and 0.1 dropout ratio.
For our proposed DARSD framework, component-specific hyperparameters are configured as follows: In the Adv-LCIB module, the learnable basis $B^{\textit{inv}}_{\textit{obs}} \in \mathbb{R}^{d \times m}$ uses $m=24$ basis vectors. The PPGCE component employs a confidence ratio $\eta$ that starts at 0.1 and increases by 0.05 every 15 training batches. In hybrid loss function $\mathcal{L}_{\textit{total}}$ (Eq.~\eqref{eq:hybrid_loss}), anti-divergence loss $\mathcal{L}_{\textit{anti}}$ and adversarial loss $\mathcal{L}_{\textit{adv}}$ are balanced with equal weights of $\lambda_1 = \lambda_2 = 0.5$.

\begin{figure}[]
    \centering
    \includegraphics[width=0.95\linewidth]{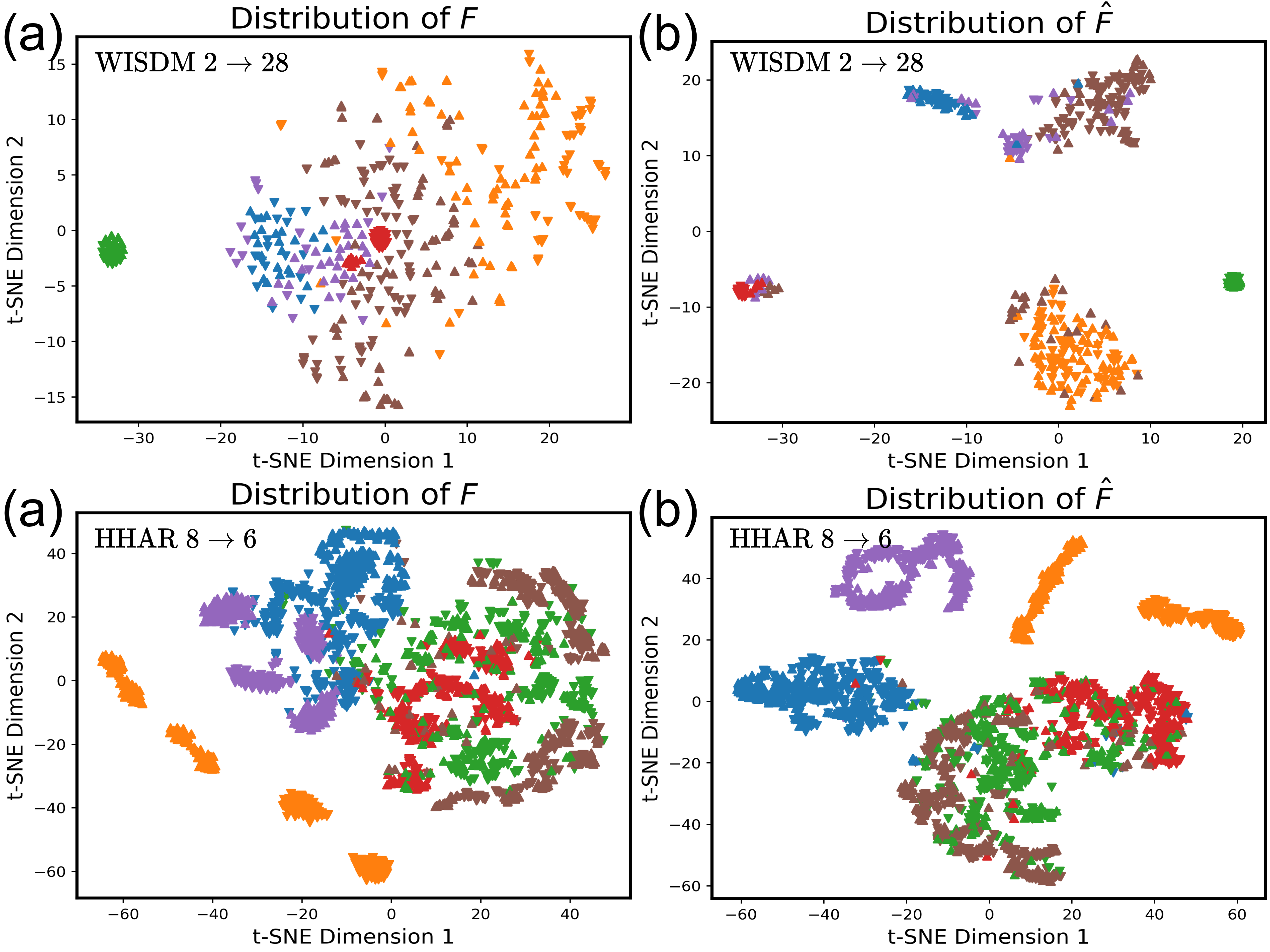}
    \caption{Distributions of $F$ and $\hat{F}$. (a) and (b) illustrate distributions of $F$ and corresponding $\hat{F}$, respectively. The \textbf{extensive examples} are displayed in \textbf{Appendix Fig. 1}.}
    \label{fig:effect_LCIB-1}
\end{figure}

\subsection{Performance Analysis}

\begin{table*}[!ht]
\centering
\small
\setlength{\tabcolsep}{3pt}
\begin{tabular}{l|ccccccccccccc}
\bottomrule

\toprule
Src $\to$ Trg    & DDC & D-CORAL & VRADA & CDAN & CAN & CoDATS & HoMM & MMDA & AdvSKM & DSAN & CLUDA & CADT & \textbf{DARSD} \\ 
\bottomrule

\toprule

\multicolumn{14}{c}{\textbf{WISDM}; \textit{Out of \textbf{19} scenarios, DARSD achieves the \textbf{15} optimal results and \underline{3} runner-up positions.}} \\ \midrule
0 $\to$ 13 & 0.319 & 0.608 & 0.441 & 0.381 & 0.319 & 0.347 & 0.577 & 0.288 & 0.563 & 0.577 & 0.409 & \underline{0.614} & \textbf{0.625} \\
1 $\to$ 24 & \underline{0.875} & 0.746 & 0.750 & 0.360 & 0.515 & 0.500 & 0.418 & 0.360 & 0.304 & 0.676 & 0.812 & 0.381 & \textbf{1.000} \\
2 $\to$ 28 & 0.669 & 0.726 & 0.688 & 0.644 & 0.610 & 0.688 & 0.691 & 0.677 & 0.742 & 0.654 & \underline{0.788} & 0.693 & \textbf{0.822} \\  
4 $\to$ 11 & 0.309 & 0.585 & 0.264 & 0.231 & 0.146 & 0.226 & 0.207 & 0.226 & 0.211 & 0.578 & \underline{0.614} & 0.448 & \textbf{0.712} \\
5 $\to$ 12 & 0.484 & 0.511 & 0.573 & 0.094 & 0.521 & 0.473 & 0.508 & 0.086 & 0.515 & 0.405 & 0.518 & \textbf{0.653} & \underline{0.577} \\
7 $\to$ 2 & 0.496 & 0.490 & 0.399 & 0.543 & 0.490 & 0.494 & 0.494 & 0.459 & 0.476 & 0.481 & 0.576 & \underline{0.632} & \textbf{0.667} \\  
7 $\to$ 26 & 0.412 & 0.396 & 0.308 & 0.344 & 0.395 & 0.405 & 0.406 & 0.385 & 0.416 & 0.401 & 0.403 & \underline{0.447} & \textbf{0.456} \\  
\midrule
\rowcolor{gray}
Avg. Rank & 7.21 & 6.84 & 7.00 & 9.89 & 8.26 & 8.16 & 6.79 & 11.2 & 7.26 & 8.58 & 4.11 & \underline{3.32} & \textbf{1.42} \\ 
\bottomrule

\toprule
\multicolumn{14}{c}{\textbf{HAR}; \textit{Out of \textbf{11} scenarios, DARSD achieves the \textbf{6} optimal results and \underline{4} runner-up positions.}} \\ \midrule
2 $\to$ 11 & 0.372 & 0.670 & 0.680 & 0.358 & 0.644 & 0.434 & 0.545 & 0.374 & 0.385 & 0.428 & 0.644 & \textbf{0.782} & \underline{0.717} \\
3 $\to$ 20 & 0.671 & 0.853 & 0.847 & 0.549 & 0.769 & 0.852 & 0.828 & 0.757 & 0.860 & 0.784 & \textbf{0.968} & 0.864 & \underline{0.907} \\
5 $\to$ 18 & 0.552 & 0.595 & 0.715 & 0.491 & 0.636 & 0.626 & \underline{0.727} & 0.462 & 0.668 & 0.663 & 0.659 & 0.710 & \textbf{0.745} \\
7 $\to$ 13 & 0.511 & 0.436 & 0.680 & 0.442 & 0.530 & 0.505 & 0.611 & 0.388 & 0.533 & 0.639 & \underline{0.884} & \textbf{0.886} & 0.681 \\
9 $\to$ 18 & 0.453 & 0.665 & \underline{0.715} & 0.357 & 0.362 & 0.389 & 0.508 & 0.319 & 0.461 & 0.590 & 0.610 & 0.653 & \textbf{0.846} \\
12 $\to$ 16 & 0.446 & 0.663 & \underline{0.689} & 0.655 & 0.500 & 0.432 & 0.678 & 0.385 & 0.446 & 0.561 & 0.582 & 0.585 & \textbf{0.832} \\
13 $\to$ 19 & 0.774 & 0.716 & 0.797 & 0.548 & 0.619 & 0.852 & 0.852 & 0.582 & 0.828 & \textbf{0.906} & \textbf{0.906} & 0.811 & \underline{0.864} \\
\midrule
\rowcolor{gray}
Avg. Rank & 10.2 & 6.55 & 5.55 & 11.0 & 8.45 & 8.18 & 4.82 & 12.0 & 8.36 & 6.36 & 3.73 & \underline{3.64} & \textbf{1.64} \\
\bottomrule

\toprule
\multicolumn{14}{c}{\textbf{HHAR}; \textit{Out of \textbf{13} scenarios, DARSD achieves the \textbf{9} optimal results and \underline{4} runner-up positions.}} \\ \midrule
0 $\to$ 2 & 0.605 & 0.569 & 0.536 & 0.611 & 0.598 & 0.598 & 0.627 & 0.612 & 0.628 & 0.415 & \textbf{0.710} & 0.585 & \underline{0.651} \\  
0 $\to$ 4 & 0.266 & 0.261 & 0.284 & 0.264 & 0.257 & 0.253 & 0.317 & 0.214 & 0.341 & 0.443 & \underline{0.589} & 0.410 & \textbf{0.619} \\
1 $\to$ 6 & 0.678 & 0.725 & 0.702 & 0.727 & 0.621 & 0.696 & 0.726 & 0.693 & 0.662 & 0.696 & \underline{0.858} & 0.782 & \textbf{0.861} \\  
2 $\to$ 3 & 0.297 & 0.437 & 0.380 & 0.338 & 0.408 & 0.444 & 0.303 & 0.253 & 0.393 & 0.598 & 0.582 & \underline{0.646} & \textbf{0.692} \\
2 $\to$ 4 & 0.231 & 0.305 & 0.415 & 0.431 & 0.294 & 0.320 & 0.230 & 0.192 & 0.219 & 0.143 & \textbf{0.526} & 0.464 & \underline{0.480} \\  
4 $\to$ 0 & 0.175 & 0.249 & 0.243 & 0.273 & 0.165 & 0.222 & 0.179 & 0.162 & 0.163 & 0.116 & \underline{0.352} & 0.304 & \textbf{0.483} \\  
4 $\to$ 1 & 0.456 & 0.461 & 0.545 & 0.667 & 0.523 & 0.469 & 0.607 & 0.517 & 0.466 & 0.488 & \underline{0.751} & 0.646 & \textbf{0.796} \\  
\midrule
\rowcolor{gray}
Avg. Rank & 9.38 & 7.31 & 6.85 & 6.77 & 7.54 & 9.00 & 6.62 & 10.7 & 9.15 & 9.00 & \underline{3.38} & 3.85 & \textbf{1.31} \\
\bottomrule

\toprule
\multicolumn{14}{c}{\textbf{MFD}; \textit{Out of \textbf{10} scenarios, DARSD achieves the \textbf{5} optimal results and \underline{2} runner-up positions.}} \\ \midrule
0 $\to$ 3 & 0.531 & 0.513 & 0.463 & 0.598 & 0.468 & 0.614 & 0.504 & \underline{0.644} & 0.493 & 0.439 & 0.605 & 0.546 & \textbf{0.716} \\
1 $\to$ 3 & 0.940 & 0.892 & 0.521 & 0.987 & 0.625 & 0.994 & 0.940 & \textbf{0.997} & 0.930 & \underline{0.996} & 0.900 & 0.765 & \textbf{0.997} \\
2 $\to$ 1 & 0.893 & 0.714 & 0.493 & \textbf{0.997} & 0.587 & \underline{0.994} & 0.849 & 0.971 & 0.887 & 0.993 & 0.559 & 0.613 & 0.953 \\
0 $\to$ 2 & 0.762 & 0.490 & 0.494 & \textbf{0.772} & 0.477 & 0.502 & 0.540 & 0.701 & 0.508 & 0.426 & 0.689 & 0.555 & \underline{0.770} \\
3 $\to$ 2 & 0.737 & 0.695 & 0.485 & \underline{0.899} & 0.564 & \textbf{0.906} & 0.729 & 0.783 & 0.728 & 0.846 & 0.790 & 0.686 & 0.813 \\ 
2 $\to$ 0 & 0.659 & 0.609 & 0.445 & \underline{0.689} & 0.489 & 0.669 & 0.613 & 0.594 & 0.562 & 0.627 & 0.628 & 0.637 & \textbf{0.792} \\
3 $\to$ 0 & 0.573 & 0.527 & 0.474 & 0.552 & 0.471 & 0.679 & 0.521 & 0.526 & 0.507 & 0.511 & 0.626 & \underline{0.736} & \textbf{0.756} \\ 
\midrule
\rowcolor{gray}
Avg. Rank & 5.70 & 8.90 & 12.4 & \underline{3.10} & 10.8 & \underline{3.10} & 7.80 & 4.80 & 9.10 & 8.50 & 7.50 & 6.90 & \textbf{2.10} \\
\bottomrule

\toprule
\end{tabular}
\caption{\textbf{Partial} Macro-F1 results on WISDM, HAR, HHAR, and MFD benchmarks. Each result is the mean over ten random initializations. Src and Trg denote the source and target subsets. The best two results are highlighted by \textbf{1st} and \underline{2nd}. The \textbf{remaining results} appear in \textbf{Appendix Table 2}.}
\label{tab:macro-f1_1}
\end{table*}

\begin{table*}[]
\centering
\small
\setlength{\tabcolsep}{5.5pt}
\begin{tabular}{cccccc|cccccccccc}
\bottomrule

\toprule
\multirow{3}{*}{ID.} & \multicolumn{5}{c}{Elements of DARSD} & \multicolumn{10}{c}{Src $\to$ Trg} \\ \cmidrule(lr){2-6} \cmidrule(lr){7-16}
 & \multirow{2}{*}{LCIB} & \multirow{2}{*}{$\mathcal{L}_{\textit{adv}}$} & \multirow{2}{*}{$\mathcal{L}_{\textit{sup}}$} & \multirow{2}{*}{$\mathcal{L}_{\textit{self}}$} & \multirow{2}{*}{$\mathcal{L}_{\textit{anti}}$} & \multicolumn{2}{c}{WISDM $12 \to 19$} & \multicolumn{2}{c}{HAR $20 \to 6$} & \multicolumn{2}{c}{HHAR $8 \to 6$} & \multicolumn{2}{c}{WISDM $2 \to 28$} & \multicolumn{2}{c}{HHAR $7 \to 4$} \\
 &  &  &  &  &  & MF1 & Acc & MF1 & Acc & MF1 & Acc & MF1 & Acc & MF1 & Acc \\ \midrule
1 &              &              & $\checkmark$ & $\checkmark$ & $\checkmark$ & 0.435 & 0.727 & 0.667 & 0.694 & 0.052 & 0.183 & 0.450 & 0.711 & 0.055 & 0.199 \\
2 & $\checkmark$ &              & $\checkmark$ & $\checkmark$ & $\checkmark$ & 0.795 & 0.838 & 0.866 & 0.878 & 0.808 & 0.803 & \underline{0.800} & 0.804 & 0.749 & 0.756 \\
3 & $\checkmark$ & $\checkmark$ & $\checkmark$ &              &              & 0.606 & 0.818 & 0.808 & 0.823 & 0.783 & 0.777 & 0.713 & 0.770 & 0.708 & 0.725 \\
4 & $\checkmark$ & $\checkmark$ & $\checkmark$ & $\checkmark$ &              & \underline{0.831} & \underline{0.853} & \underline{0.903} & \underline{0.912} & \textbf{0.862} & \textbf{0.857} & 0.783 & \underline{0.815} & \underline{0.782} & \underline{0.789} \\
\rowcolor{gray}
5 & $\checkmark$ & $\checkmark$ & $\checkmark$ & $\checkmark$ & $\checkmark$ & \textbf{0.883} & \textbf{0.917} & \textbf{0.967} & \textbf{0.989} & \underline{0.844} & \underline{0.848} & \textbf{0.822} & \textbf{0.883} & \textbf{0.813} & \textbf{0.817} \\ 
\bottomrule

\toprule
\end{tabular}
\caption{Ablation analysis of DARSD\@. Specifically, the contributions of LCIB, $\mathcal{L}_{\textit{adv}}$, $\mathcal{L}_{\textit{self}}$, and $\mathcal{L}_{\textit{anti}}$ are verified. The check mark ($\checkmark$) denotes the activated components. We \textbf{bold} the best results and \underline{underlined} the runner-up positions.}
\label{tab:ablation}
\end{table*}

We compare DARSD with 12 UDA baselines on the WISDM, HAR, HHAR, and MFD benchmarks in Table~\ref{tab:macro-f1_1}. 

\textbf{DARSD achieves optimal performance in 35 out of 53 scenarios and ranks first across all datasets}, demonstrating consistent superiority.
On WISDM and HHAR, DARSD achieves an exceptional average rank of 1.42 and 1.31, substantially outperforming CADT (3.32) and CLUDA (3.38), respectively. This robust performance despite class imbalance demonstrates DARSD's ability to effectively disentangle domain-invariant action modes information about from individual-specific artifacts. 
On MFD, a dataset fundamentally distinct from human activity benchmarks ($i.e.$, WISDM, HAR, and HHAR) due to its dense high-frequency and long-term industrial vibration signals, DARSD maintains competitive performance with an average rank of 2.10 compared to CDAN and CoDATS, both achieving an average rank of 3.10. 
This cross-scenario generalization from personal sensing to industrial monitoring validates the broad applicability of our explicit decomposition approach.

The comparative analysis illuminates advantages of our DARSD over existing paradigms. Adversarial methods ($e.g.$, CDAN and VRADA) exhibit inconsistent performance, as they risk eliminating not only domain-specific artifacts while also also valid semantic information. Metric learning approaches ($e.g.$, HoMM and CADT) demonstrate moderate consistency. While the self-supervision method CLUDA performs strongly, relying on gradually and slowly contrastive learning and adversarial training. In contrast, our DARSD framework provides more direct control over trade-off between invariance and discriminability, with faster convergence. The \textbf{\textit{Convergence Efficiency}} of DARSD and CLUDA is compared in \textbf{Appendix D.4}.
To better characterize performance of DARSD, we conduct \textbf{\textit{Representation Visualization}} of source and target features in \textbf{Appendix D.3}.

\subsection{Ablation Study}
To validate the contribution of each component, we conduct comprehensive ablation studies across five randomly selected scenarios, progressively adding modules to demonstrate their individual and synergistic effects in Table~\ref{tab:ablation}. 

\textbf{Impact of Explicit Decomposition (LCIB)}: 
The comparison between ID.1 and ID.5 (full model) underscores the fundamental importance of explicit representation decomposition. Without LCIB module, performance degrades dramatically across all scenarios. 
This striking contrast validates the effectiveness of LCIB module in extracting semantic domain-invariant patterns.
Consistent with Fig.~\ref{fig:effect_LCIB-1}, reconstructed domain-invariant features $\hat{F}$ in Fig.~\ref{fig:effect_LCIB-1}b exhibit a more orderly distribution compared to those in Fig.~\ref{fig:effect_LCIB-1}a.

\textbf{Necessity of Adversarial Training ($\mathcal{L}_{\textit{adv}}$)}: 
Comparing ID.2 and ID.5 reveals that adversarial training is crucial for ensuring the quality of $B^{\textit{inv}}_{\textit{obs}}$. 
Consistent improvements across all scenarios demonstrate that without adversarial constraints, LCIB may inadvertently capture domain-specific information, undermining the invariance property. 
This confirms the rationality of adversarial training of $B^{\textit{inv}}_{\textit{obs}}$.

\textbf{Value of Target Feature Utilization ($\mathcal{L}_{\textit{self}}$ and $\mathcal{L}_{\textit{anti}}$)}: 
The progression from ID.3 to ID.4 and then to ID.5 illustrates the importance of effective utilization of all $\hat{F}_t$. ID.3's performance shows that relying solely on $\hat{F}^{\textit{con}}_t$ leaves significant room for improvement. Incorporating self-supervised consistency (ID.4) yields notable gains, while anti-divergence regularization (ID.5) provides the final performance boost by bridging $\{ \hat{F}_s \cup \hat{F}^{\textit{con}}_t \}$ and $\hat{F}^{\textit{dis}}_t$.

\textbf{Synergistic Effects}: 
The significant performance gap between any individual component and the full DARSD model demonstrates that its effectiveness arises from component synergy, not mere aggregation. 
\\

\noindent
\textbf{Notations: Due to space constraints, A. Notation, B. Proof of Representation Space Decomposition, C. Detailed Benchmark Description, D.1. Macro-F1 Results, D.2. Hyperparameter Sensitivity Analysis, D.3. Representation Visualization, D.4. Convergence Efficiency, and E. Related Works are provided in Appendix.}

\section{Conclusions}
\label{sec:conclusion}
In this paper, we \textit{for the first time} shift the paradigm of unsupervised time series domain adaptation from the perspective of explicit \textbf{representation space decomposition}. 
Building on this insight, we propose \textbf{DARSD}, disentangling domain-invariant patterns from domain-specific artifacts via a learnable invariant subspace basis. Additionally, DARSD decouples the optimization objectives of invariant feature extraction and robust cross-domain feature discriminative aggregation through Adv-LCIB and hybrid contrastive optimization with PPGCE modules. 
Extensive experiments conducted on \textbf{four benchmarks} demonstrate its superiority. Comprehensive studies prove the rationality of individual elements in DARSD and their synergy to achieve optimal performance.
Future work will explore adaptive basis selection to further enhance the DARSD's applicability across diverse time series domains.

\bibliography{aaai2026}

\newpage

\appendix

\renewcommand{\tablename}{Appendix Table}
\renewcommand{\figurename}{Appendix Figure}

\setcounter{figure}{0}
\setcounter{table}{0}
\setcounter{equation}{0}

\begin{strip}
    \centering
    \LARGE\bfseries Appendix \\[1ex]
    \par
\end{strip}

\section{A. Notation}
For easier reading, the crucial notations involved in this manuscript are summarized in Appendix Table~\ref{tab:notation}.

\begin{table}[!ht]
  \centering
  \small
  \begin{tabular}{c|c}
  \bottomrule
  
  \toprule
  Symbol Formula                                                                                                & Definition                                            \\ \midrule
  $x^s_i, \; x^t_j \in \mathbb{R}^{T \times D}$                                                                 & Source and target samples                             \\ \midrule
  $\mathcal{S} = \{ (x^s_i, y^s_i) \}^{n_s}_{i=1} \sim \mathcal{D}_s$                          & \makecell{Source domain datasets \\ following source distribution $\mathcal{D}_s$}         \\ \midrule
  $\mathcal{T} = \{ x^t_i \}^{n_t}_{i=1} \sim \mathcal{D}_t$                                   & \makecell{Target domain datasets \\ following target distribution $\mathcal{D}_t$}         \\ \midrule 
  $B_{\textit{obs}}^{\textit{inv}} \in \mathbb{R}^{d \times m}$                                                 & Learnable common invariant basis                      \\ \midrule
  $w^{\textit{inv}} \in \mathbb{R}^{m}$                                                        & \makecell{Coordinate of  feature \\ in domain-invariant subspace}                          \\ \midrule
  $\hat{w} \in \mathbb{R}^{m}$                                                                 &  Softmax regularized coordinate                            \\ \midrule
  $\hat{f}^s_i, \; \hat{f}^t_i \in \mathbb{R}^{d}$                                                              & \makecell{Reconstructed domain-invariant \\ features}               \\ \midrule
  $\hat{F}_s = \{ \hat{f}^s_i \}_{i=1}^{n_s} $                                                                  & Set of reconstructed source features                  \\ \midrule
  $\eta(t) \in [0, 1]$                                                                                          & Confidence ratio for $\hat{F}_t$                      \\ \midrule
  $\sigma_i $                                                                                                   & Confidence score of $\hat{f}^t_i$      \\ \midrule
  $\hat{f}^{\textit{con}}_i$                                                      & Confident target feature               \\ 
  \midrule
  $\hat{f}^{\textit{dis}}_i $                                                      & Distrusted target feature               \\ 
  \midrule
  $\hat{F}^{\textit{con}}_t = \{ \hat{f}^{\textit{con}}_i \}^{\eta(t) n_t}_{i=1} $                              & Set of confident target features                      \\ \midrule
  $\hat{F}^{\textit{dis}}_t = \{ \hat{f}^{\textit{dis}}_i \}^{(1 - \eta(t)) n_t}_{i=1} $                        & Set of distrusted target features                     \\ \midrule
  $\hat{F}_t = \{ \hat{F}_t^{\textit{con}} \cup \hat{F}_t^{\textit{dis}} \}$    & Set of reconstructed target features                  \\
  \bottomrule
  
  \toprule
  \end{tabular}
  \caption{Specific description of crucial notations.}
  \label{tab:notation}
\end{table}

\begin{figure}[t]
    \centering
    \includegraphics[width=1.0\linewidth]{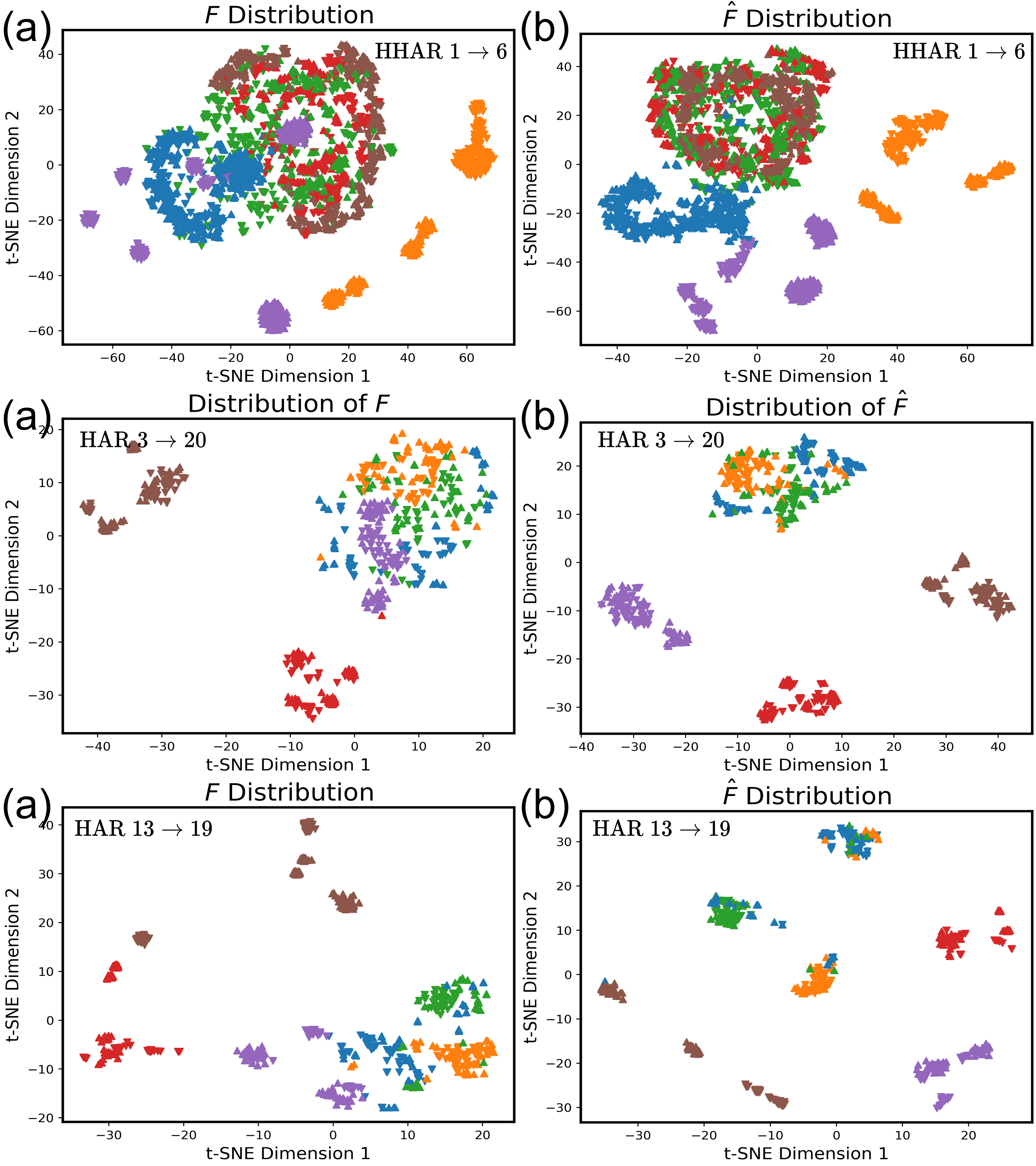}
    \caption{\textbf{Extension of Figure 3 in main text.} Distributions of $F$ and $\hat{F}$. (a) and (b) illustrate distributions of $F$ and corresponding $\hat{F}$, respectively.}
    \label{app_fig:effect_LCIB}
\end{figure}

\section{B. Proof of Representation Space Decomposition}\label{app:proof}

\subsection{B.1. Problem Setup}
Consider a feature vector $f \in \mathbb{R}^{d}$ extracted from sample $x$. We decompose $f$ into two orthogonal components:
\begin{equation}
    f = f^{\textit{inv}} + f^{\textit{spe}}
\end{equation}
where $f^{\textit{inv}}$ and $f^{\textit{spe}}$ represent domain-invariant patterns and domain-specific noise, respectively.

\subsection{B.2. Subspace Representation}
We assume that $\mathbb{R}^d$ can be decomposed into two orthogonal subspaces:
\begin{equation}
    \mathbb{R}^{d} = \mathcal{S}^{\textit{inv}} \oplus \mathcal{S}^{\textit{spe}}
\end{equation}
where $\mathcal{S}^{\textit{inv}}$ and $\mathcal{S}^{\textit{spe}}$ are the domain-invariant and domain-specific subspaces with dimensions $m$ and $\bar{m} = d - m$, respectively.

Let $B^{\textit{inv}} \in \mathbb{R}^{d \times m}$ and $B^{\textit{spe}} \in \mathbb{R}^{d \times \bar{m}}$ be orthonormal basis matrices for these subspaces such that:
\begin{gather}
\begin{aligned}
    (B^{\textit{inv}})^{\top} B^{\textit{inv}} &= \mathbf{I}_{m},    
    \\
    (B^{\textit{spe}})^{\top} B^{\textit{spe}} &= \mathbf{I}_{\bar{m}},    
    \\
    (B^{\textit{inv}})^{\top} B^{\textit{spe}} &= \mathbf{0}_{m \times \bar{m}}
\end{aligned}
\end{gather}

Then the decomposed components can be expressed as:
\begin{gather}
    f^{\textit{inv}} = B^{\textit{inv}} w^{\textit{inv}} 
    \tag{6}
    \\
    f^{\textit{spe}} = B^{\textit{spe}} w^{\textit{spe}} 
    \tag{7}
\end{gather}
where $w^{\textit{inv}} \in \mathbb{R}^{m}$ and $w^{\textit{spe}} \in \mathbb{R}^{\bar{m}}$ are the coordinate vectors in their respective subspaces.

\subsection{B.3. Domain-Invariant Component Extraction}
\textbf{Assumption}: We have access to an observable approximation $B^{\textit{inv}}_{\textit{obs}} \in \mathbb{R}^{d \times m}$ of the true domain-invariant basis $B^{\textit{inv}}$ such that:
\begin{equation}
    (B^{\textit{inv}})^{\top} B^{\textit{inv}}_{\textit{obs}} = \mathbf{I}_m
\end{equation}

This assumption implies that $B^{\textit{inv}}_{\textit{obs}}$ spans the same subspace as $B^{\textit{inv}}$ and maintains orthogonality with $B^{\textit{spe}}$.

\textbf{Theorem}: The domain-invariant coordinates can be extracted via projection:
\begin{equation}
    w^{\textit{inv}} = (B^{\textit{inv}}_{\textit{obs}})^{\top} f
\end{equation}

\textbf{Proof}:
\begin{equation}
\begin{aligned}
    (B^{\textit{inv}}_{\textit{obs}})^{\top} f &= (B^{\textit{inv}}_{\textit{obs}})^{\top} (f^{\textit{inv}} + f^{\textit{spe}}) \\
    &= (B^{\textit{inv}}_{\textit{obs}})^{\top} (B^{\textit{inv}} w^{\textit{inv}} + B^{\textit{spe}} w^{\textit{spe}}) \\
    &= (B^{\textit{inv}}_{\textit{obs}})^{\top} B^{\textit{inv}} w^{\textit{inv}} + (B^{\textit{inv}}_{\textit{obs}})^{\top} B^{\textit{spe}} w^{\textit{spe}} \\
    &= \mathbf{I}_m w^{\textit{inv}} + \mathbf{0}_{m \times \bar{m}} w^{\textit{spe}} \\
    &= w^{\textit{inv}}
\end{aligned}
\end{equation}

The key steps utilize:

1. Equation (4): $(B^{\textit{inv}})^{\top} B^{\textit{inv}}_{\textit{obs}} = \mathbf{I}_m$, which implies $(B^{\textit{inv}}_{\textit{obs}})^{\top} B^{\textit{inv}} = \mathbf{I}_m$.

2. Orthogonality: $(B^{\textit{inv}}_{\textit{obs}})^{\top} B^{\textit{spe}} = \mathbf{0}_{m \times \bar{m}}$.

\subsection{B.4. Domain-Invariant Feature Reconstruction}

Once we obtain coordinate $w^{\textit{inv}}$, we can reconstruct the domain-invariant component:
\begin{equation}
    f^{\textit{inv}} = B^{\textit{inv}}_{\textit{obs}} w^{\textit{inv}}
\end{equation}

\section{C. Detailed Benchmark Description}\label{app:dataset}
The detailed backgrounds of WISDM, HAR, HHAR, and MFD benchmarks are displayed as follows:
\begin{itemize}
    \item Wireless Sensor Data Mining (\textbf{WISDM}) Dataset \cite{10.1145/1964897.1964918}: This dataset comprises 3-axis accelerometer measurements collected from 30 participants, sampled at a frequency of 20 Hz. Each measurement is segmented into non-overlapping segments of 128-time steps to predict the activity label for each participant. The dataset covers six activity labels: walking, jogging, sitting, standing, walking upstairs, and walking downstairs. Notably, the dataset poses challenges due to its imbalanced class distribution.
    \item Human Activity Recognition (\textbf{HAR}) Dataset~\cite{Anguita2013APD}: This dataset features measurements from a 3-axis accelerometer, 3-axis gyroscope, and 3-axis body acceleration collected from 30 participants at a sampling rate of 50 Hz. The data is segmented into non-overlapping segments of 128-time steps for action classification. We classify the time series data into six types of activities: walking, walking upstairs, walking downstairs, sitting, standing, and lying down.
    \item Heterogeneity Human Activity Recognition (\textbf{HHAR}) Dataset~\cite{stisen2015smart}: This dataset comprises 3-axis accelerometer measurements from 30 participants, sampled at 50 Hz. Segmented into non-overlapping 128-time step sequences, the dataset aims to classify activities across six labels: biking, sitting, standing, walking, walking upstairs, and walking downstairs.
    \item Machine Fault Diagnosis (\textbf{MFD}) Dataset~\cite{lessmeier2016condition}: This dataset was collected by Paderborn University for bearing fault diagnosis, utilizing vibration signals to identify various types of incipient faults. The data were gathered under four distinct working conditions. Each sample consists of a single univariate channel containing 5,120 data points.
\end{itemize}

\begin{figure}[t]
    \centering
    \subfloat[Hyperparameter $m$.]{
        \includegraphics[width=0.496\linewidth]{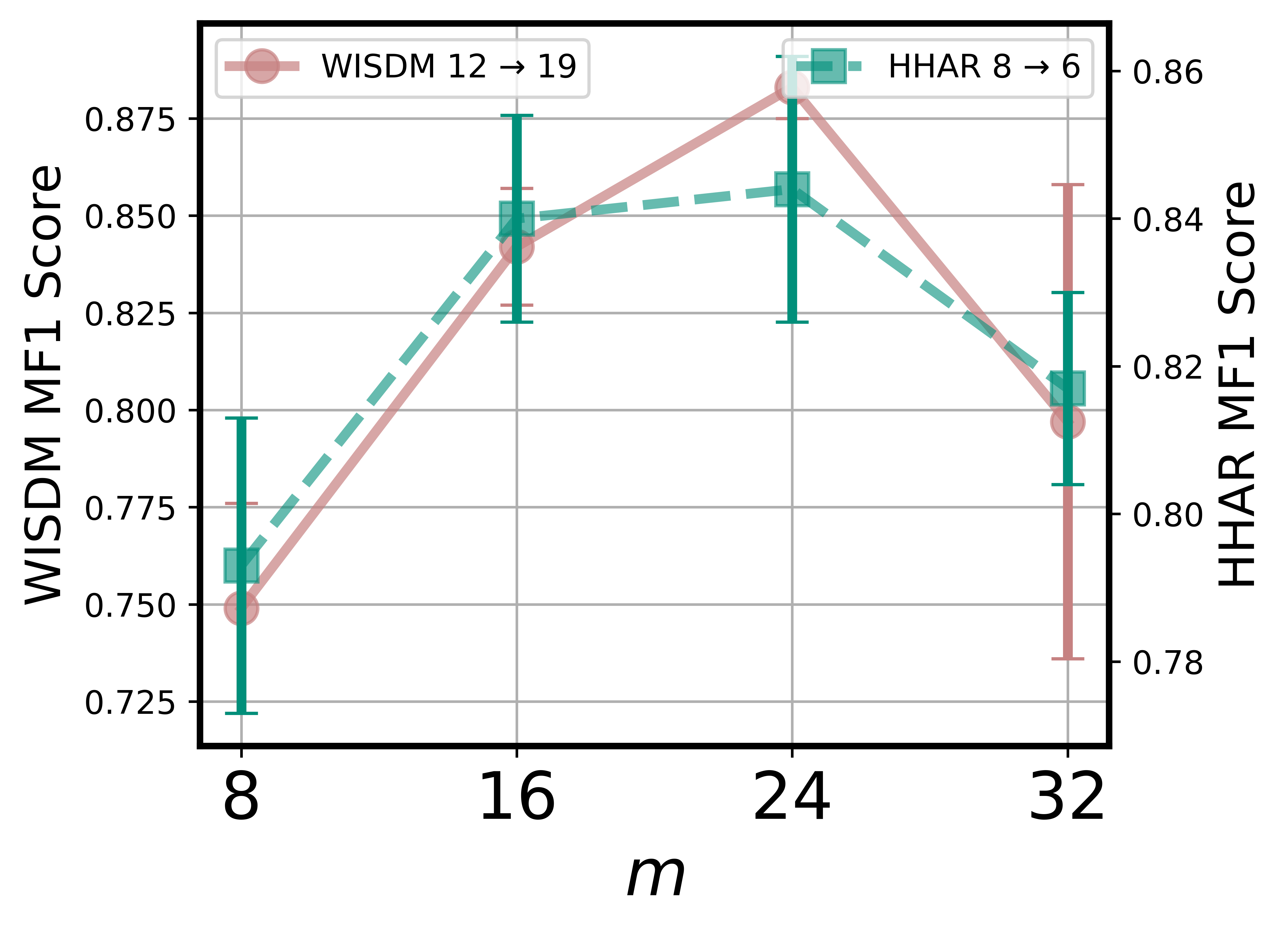}
        \label{subfig:rob_exp_1}}
    \subfloat[Hyperparameter $\lambda_2$.]{
        \includegraphics[width=0.496\linewidth]{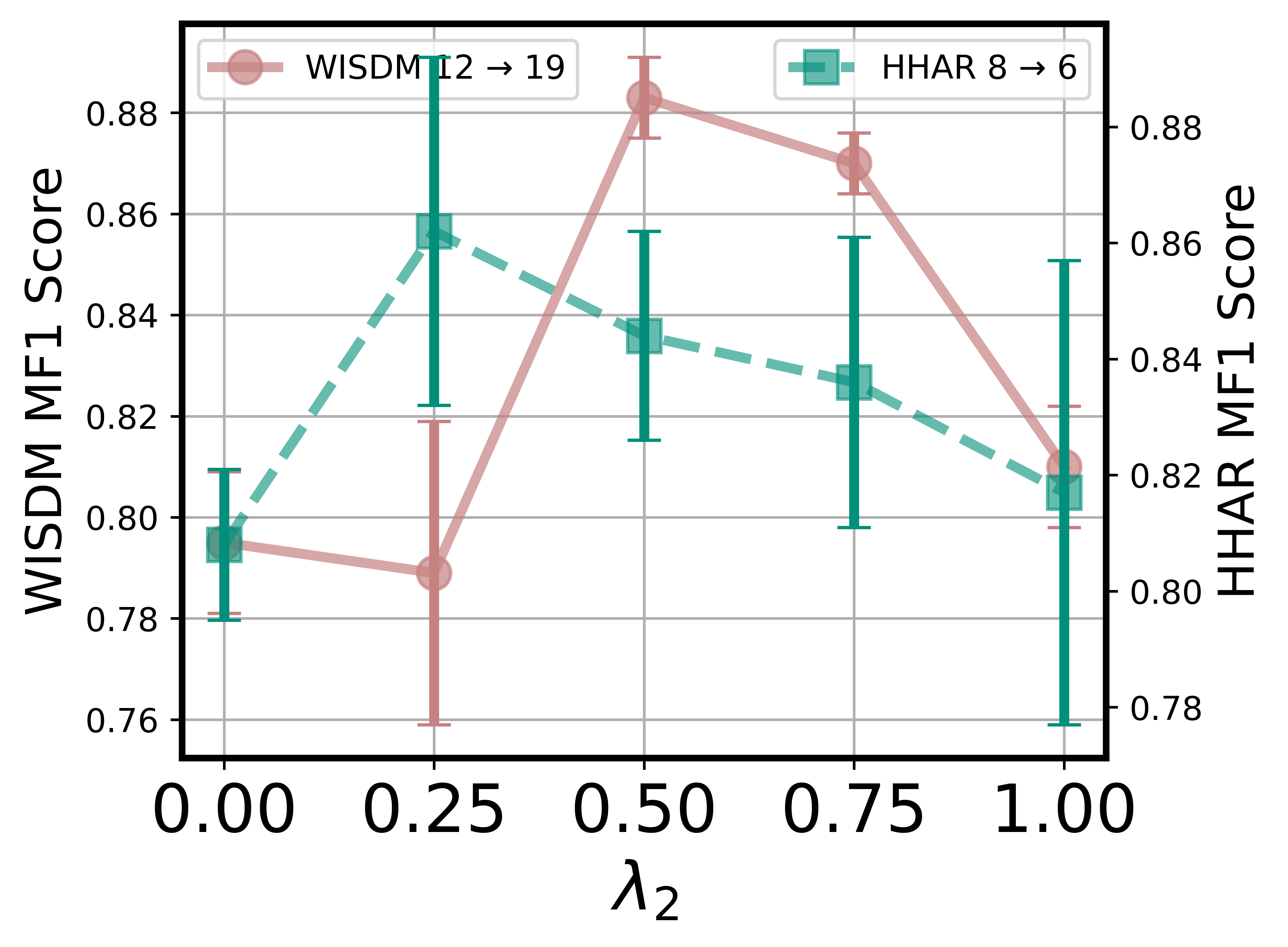}
        \label{subfig:rob_exp_2}}
    \caption{Hyperparameter sensitivity analysis of $m$ and $\lambda_2$ on WISDM 12 $\to$ 19 and HHAR 8 $\to$ 6 scenarios.}
    \label{fig:rob_exp}
\end{figure}

\section{D. More Experimental Results}

\subsection{D.1. Macro-F1 Results}

\begin{table*}[!ht]
\centering
\small
\setlength{\tabcolsep}{3pt}
\begin{tabular}{l|ccccccccccccc}
\bottomrule

\toprule
Src $\to$ Trg    & DDC & D-CORAL & VRADA & CDAN & CAN & CoDATS & HoMM & MMDA & AdvSKM & DSAN & CLUDA & CADT & \textbf{DARSD} \\ 
\bottomrule

\toprule

\multicolumn{14}{c}{\textbf{WISDM}; \textit{Out of \textbf{19} scenarios, DARSD achieves the \textbf{15} optimal results and \underline{3} runner-up positions.}} \\ \midrule
10 $\to$ 14 & 0.541 & 0.337 & 0.591 & 0.266 & \underline{0.608} & 0.477 & 0.667 & 0.469 & 0.471 & 0.337 & 0.510 & 0.599 & \textbf{0.691} \\
10 $\to$ 21 & \textbf{0.694} & 0.250 & 0.125 & 0.118 & 0.250 & 0.390 & 0.313 & 0.390 & \textbf{0.694} & 0.106 & 0.250 & \underline{0.518} & 0.392 \\
12 $\to$ 7 & 0.632 & 0.486 & 0.437 & 0.546 & 0.636 & 0.612 & 0.442 & 0.539 & 0.655 & 0.574 & \underline{0.678} & 0.620 & \textbf{0.834} \\  
12 $\to$ 19 & 0.396 & 0.317 & 0.410 & 0.298 & 0.508 & 0.456 & 0.281 & 0.233 & 0.510 & 0.518 & \underline{0.532} & 0.518 & \textbf{0.883} \\  
15 $\to$ 29 & 0.299 & 0.265 & 0.391 & 0.238 & 0.372 & 0.136 & 0.208 & 0.133 & 0.299 & 0.182 & \underline{0.482} & 0.441 & \textbf{0.529} \\
18 $\to$ 20 & 0.383 & 0.379 & 0.578 & 0.600 & 0.389 & 0.427 & 0.421 & 0.280 & 0.348 & 0.268 & 0.673 & \underline{0.702} & \textbf{0.658} \\  
19 $\to$ 2 & 0.459 & 0.501 & \underline{0.615} & 0.312 & 0.327 & 0.403 & 0.522 & 0.306 & 0.460 & 0.428 & 0.458 & 0.555 & \textbf{0.664} \\  
21 $\to$ 16 & 0.537 & 0.669 & 0.542 & 0.404 & 0.641 & 0.410 & 0.669 & 0.310 & 0.523 & 0.568 & 0.595 & \textbf{0.850} & \underline{0.787} \\
25 $\to$ 22 & 0.273 & 0.241 & \underline{0.506} & 0.401 & 0.241 & 0.200 & 0.307 & 0.159 & 0.309 & 0.341 & 0.375 & 0.500 & \textbf{0.610} \\
26 $\to$ 2 & 0.414 & 0.618 & 0.517 & 0.404 & 0.362 & 0.598 & 0.519 & 0.453 & 0.463 & 0.424 & \underline{0.701} & 0.610 & \textbf{0.731} \\  
28 $\to$ 2 & 0.484 & 0.495 & 0.473 & 0.400 & 0.412 & 0.492 & 0.511 & 0.430 & 0.484 & 0.451 & \textbf{0.710} & 0.606 & \underline{0.656} \\  
28 $\to$ 20 & 0.571 & 0.620 & 0.672 & 0.605 & 0.655 & 0.578 & 0.699 & 0.537 & 0.557 & 0.615 & 0.703 & \underline{0.726} & \textbf{0.728} \\  
\midrule
\rowcolor{gray}
Avg. Rank & 7.21 & 6.84 & 7.00 & 9.89 & 8.26 & 8.16 & 6.79 & 11.2 & 7.26 & 8.58 & 4.11 & \underline{3.32} & \textbf{1.42} \\ 
\bottomrule

\toprule
\multicolumn{14}{c}{\textbf{HAR}; \textit{Out of \textbf{11} scenarios, DARSD achieves the \textbf{6} optimal results and \underline{4} runner-up positions.}} \\ \midrule
18 $\to$ 21 & 0.439 & 0.760 & 0.697 & 0.450 & 0.449 & 0.857 & 0.803 & 0.380 & 0.439 & 0.588 & 0.920 & \underline{0.929} & \textbf{0.948} \\
20 $\to$ 6 & 0.643 & 0.776 & 0.619 & 0.407 & 0.593 & 0.659 & 0.935 & 0.407 & 0.586 & 0.588 & \underline{0.959} & 0.782 & \textbf{0.967} \\
23 $\to$ 13 & 0.398 & 0.586 & 0.683 & 0.298 & 0.509 & 0.439 & 0.539 & 0.518 & 0.401 & 0.723 & 0.720 & \underline{0.743} & \textbf{0.764} \\
24 $\to$ 12 & 0.724 & 0.765 & 0.658 & 0.758 & 0.872 & 0.709 & \textbf{0.893} & 0.448 & 0.736 & 0.826 & 0.866 & 0.768 & \underline{0.887} \\ 
\midrule
\rowcolor{gray}
Avg. Rank & 10.2 & 6.55 & 5.55 & 11.0 & 8.45 & 8.18 & 4.82 & 12.0 & 8.36 & 6.36 & 3.73 & \underline{3.64} & \textbf{1.64} \\
\bottomrule

\toprule
\multicolumn{14}{c}{\textbf{HHAR}; \textit{Out of \textbf{13} scenarios, DARSD achieves the \textbf{9} optimal results and \underline{4} runner-up positions.}} \\ \midrule
5 $\to$ 2 & 0.311 & 0.330 & 0.327 & 0.324 & \underline{0.338} & 0.209 & 0.335 & 0.278 & 0.229 & 0.232 & 0.299 & 0.318 & \textbf{0.340} \\
7 $\to$ 0 & 0.194 & 0.198 & 0.169 & \underline{0.454} & 0.178 & 0.282 & 0.232 & 0.095 & 0.189 & 0.247 & 0.438 & 0.304 & \textbf{0.527} \\
7 $\to$ 2 & 0.233 & 0.295 & \textbf{0.525} & 0.333 & 0.250 & 0.217 & 0.258 & 0.235 & 0.203 & 0.179 & 0.332 & 0.390 & \underline{0.392} \\
7 $\to$ 4 & 0.577 & 0.608 & 0.658 & 0.344 & 0.736 & 0.489 & 0.750 & 0.539 & 0.543 & 0.522 & 0.604 & \textbf{0.850} & \underline{0.813} \\
7 $\to$ 5 & 0.488 & 0.496 & 0.529 & 0.480 & 0.546 & 0.374 & 0.323 & 0.283 & 0.487 & \underline{0.649} & 0.626 & 0.645 & \textbf{0.680} \\  
8 $\to$ 6 & 0.568 & 0.728 & 0.723 & 0.334 & 0.765 & 0.701 & 0.745 & 0.514 & 0.734 & 0.467 & \underline{0.837} & 0.773 & \textbf{0.844} \\ 
\midrule
\rowcolor{gray}
Avg. Rank & 9.38 & 7.31 & 6.85 & 6.77 & 7.54 & 9.00 & 6.62 & 10.7 & 9.15 & 9.00 & \underline{3.38} & 3.85 & \textbf{1.31} \\
\bottomrule

\toprule
\multicolumn{14}{c}{\textbf{MFD}; \textit{Out of \textbf{10} scenarios, DARSD achieves the \textbf{5} optimal results and \underline{2} runner-up positions.}} \\ \midrule
1 $\to$ 0 & 0.500 & 0.514 & 0.427 & 0.555 & 0.507 & 0.614 & 0.506 & 0.515 & 0.500 & 0.484 & 0.494 & \textbf{0.709} & \underline{0.663} \\
1 $\to$ 2 & 0.749 & 0.704 & 0.500 & \underline{0.881} & 0.572 & \textbf{0.919} & 0.747 & 0.773 & 0.734 & 0.732 & 0.574 & 0.680 & 0.776 \\
0 $\to$ 1 & 0.524 & 0.504 & 0.455 & 0.599 & 0.518 & \underline{0.607} & 0.509 & 0.593 & 0.468 & 0.374 & 0.510 & 0.515 & \textbf{0.722} \\
\midrule
\rowcolor{gray}
Avg. Rank & 5.70 & 8.90 & 12.4 & \underline{3.10} & 10.8 & \underline{3.10} & 7.80 & 4.80 & 9.10 & 8.50 & 7.50 & 6.90 & \textbf{2.10} \\
\bottomrule

\toprule
\end{tabular}
\caption{\textbf{Extension of Table 2 in main text.} Macro-F1 results on WISDM, HAR, HHAR, and MFD benchmarks. Each result is the mean over ten random initializations. Src and Trg denote the source and target subsets. The best two results are highlighted by \textbf{1st} and \underline{2nd}.}
\label{tab:macro-f1_2}
\end{table*}

The Appendix Table 2 displays the remaining part of Table 2 in main text.

\subsection{D.2. Hyperparameter Sensitivity Analysis}
We analyze the sensitivity of two critical hyperparameters that govern our explicit decomposition approach: the dimensionality of the invariant basis $m$ and the weight of adversarial training $\lambda_2$.

\textbf{Invariant Basis Dimensionality ($m$)}: 
Appendix Fig.~\ref{subfig:rob_exp_1} shows that optimal performance occurs at $m = 24$ for feature dimensions $d = 128$. This indicates that a 24-dimensional subspace provides sufficient capacity to capture essential semantic domain-invariant patterns, enabling effective domain adaptation. 
When $m < 24$, the invariant subspace $B^{\textit{inv}}_{\textit{obs}}$ lacks sufficient capacity to capture semantic complexity, resulting in underfitting. 
When $m > 24$, the enlarged basis becomes harder to train effectively and more susceptible to incorporating domain-specific information, thereby undermining the invariance property.

\textbf{Adversarial Training Weight ($\lambda_2$)}: 
Appendix Fig.~\ref{subfig:rob_exp_2} reveals stable performance within $\lambda_2 \in [0.25, 0.50]$, indicating robust optimization dynamics. 
When $\lambda_2 < 0.25$, insufficient adversarial pressure results in lossy reconstruction of domain-invariant information, causing significant information loss. 
When $\lambda_2 > 0.50$, excessive adversarial pressure forces the model to incorporate domain-specific information to maintain reconstruction consistency, compromising the invariance property.

Based on the sensitivity analysis, we recommend $m = 0.2d$ and $\lambda_2 = 0.5$ for general deployment. The consistent optimal ranges across different cross-domain scenarios demonstrate the robustness of DARSD's explicit representation space decomposition approach.

\subsection{D.3. Representation Visualization}

\begin{figure*}[t]
    \centering
    \includegraphics[width=1.0\linewidth]{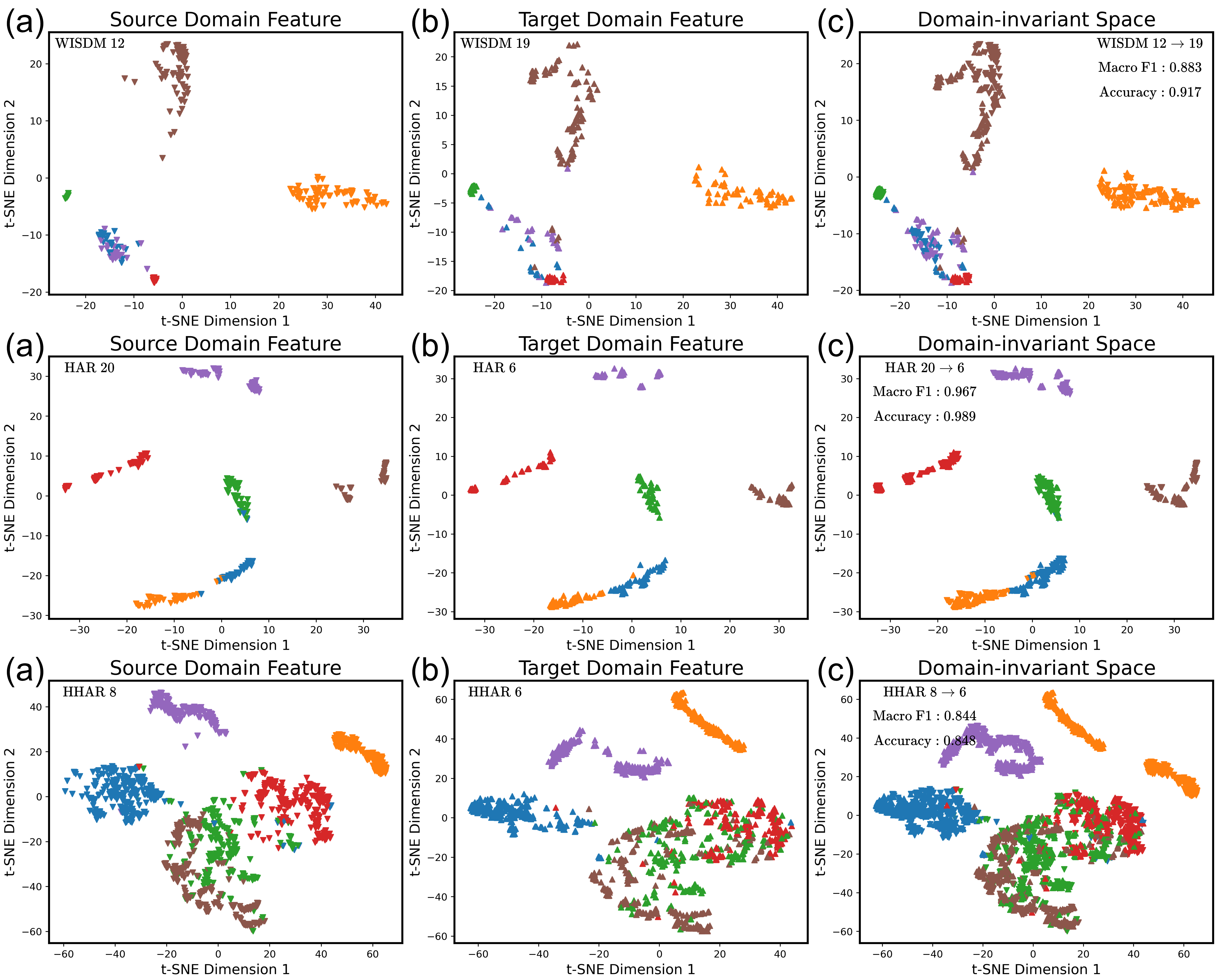}
    \caption{t-SNE visualization of DARSD feature distribution in the domain-invariant subspace. Colors distinguish different classes; downward ($\bigtriangledown$) and upward ($\bigtriangleup$) triangles denote source and target features, respectively. Each row shows: (a) source features, (b) target features, and (c) combined source-target feature distribution in the domain-invariant subspace.}
    \label{app_fig:feat_dis}
\end{figure*}

Appendix Fig.~\ref{app_fig:feat_dis} visualizes distributions of source and target features in the invariant subspace using t-SNE, providing compelling evidence for DARSD's effectiveness. 
Source features (Appendix Fig.~\ref{app_fig:feat_dis}a) form well-separated semantic clusters with distinct inter-class boundaries. Target features (Appendix Fig.~\ref{app_fig:feat_dis}b) display a remarkably similar geometric structure, where cluster shapes and relative positions mirror those of the source distribution. This structural preservation indicates that our Adv-LCIB module successfully extracts transferable domain-invariant semantic patterns while filtering domain-specific artifacts.
Appendix Fig.~\ref{app_fig:feat_dis}c displays the substantial spatial overlap between source and target features of identical classes, alongside maintained inter-class separation. This confirms that our DARSD framework achieves dual objectives of domain invariance and semantic discriminability. 
Notably, explainable feature visualization validates that learned representations are truly domain-agnostic, supporting our theoretical representation space decomposition framework and the effectiveness of PPGCE in the domain-invariant subspace.

\subsection{D.4. Convergence Efficiency}

\begin{figure}[t]
    \centering
    \includegraphics[width=1.0\linewidth]{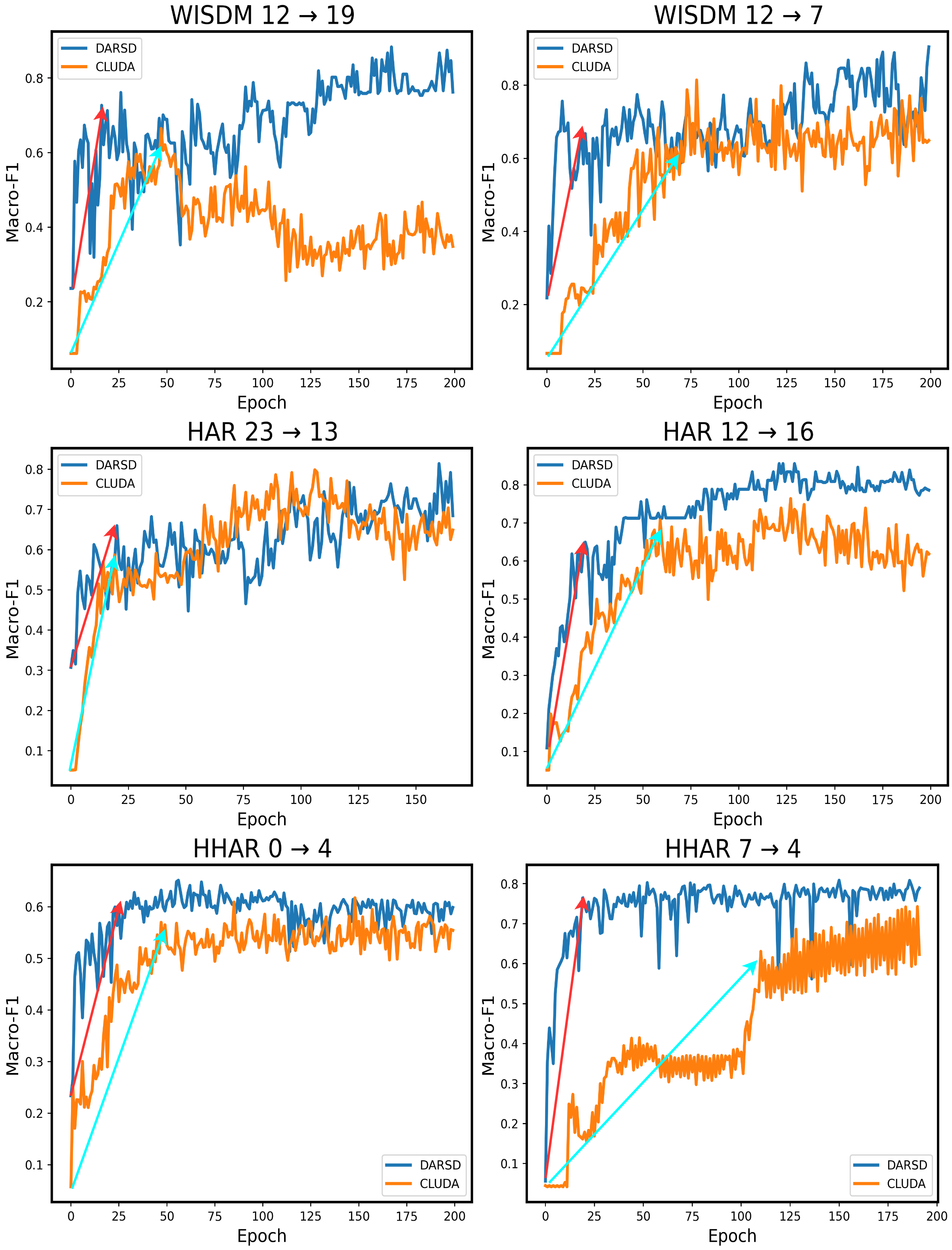}
    \caption{Comparisons of convergence efficiency of DARSD and CLUDA\@. The red and blue arrows indicate convergence trends of DARSD and CLUDA, respectively.}
    \label{fig:convergence}
\end{figure}

Appendix Fig.~\ref{fig:convergence} compares the convergence efficiency of DARSD and CLUDA across representative scenarios. DARSD consistently exhibits faster convergence, typically reaching stable performance within 50-80 epochs compared to CLUDA's 100-150 epochs.

The efficiency advantage stems from separation strategies of optimization objectives of DARSD: the Adv-LCIB module explicitly constructs domain-invariant features, while hybrid contrastive optimization directly optimizes for discrimination in the invariant subspace. This reduces optimization complexity compared to CLUDA, which simultaneously extracts domain-invariant representations and learns classification boundaries through a shared feature extractor, relying on indirect signals.

\section{E. Related Works}

\subsection{E.1. Unsupervised Domain Adaptation}
Unsupervised domain adaptation (UDA) seeks to transfer knowledge from an annotated source domain to an unlabeled target domain, where two domains have related yet different distributions. Despite significant progress, existing approaches suffer from a fundamental limitation: they treat representations as \textbf{monolithic entities} and strive to align \textbf{entire feature distributions}, failing to recognize that only part of components contain transferable knowledge. They can be categorized into three groups:
\textbf{(1) \textit{Adversarial Training Methods}} represent the earliest paradigm for UDA~\cite{tzeng2017adversarial}, employing domain discriminators to learn domain-confusing representations. MBAN~\cite{10688076} and CPDIC~\cite{10497695} utilize adversarial training to align marginal distributions, while CoDATS~\cite{wilson2020multi} extends this framework to time series with multi-source adaptation. However, these methods suffer from a critical flaw: \textit{the adversarial game fails to distinguish between transferable semantic content and domain-specific noise}, resulting in the elimination of essential information required for accurate classification.
\textbf{(2) \textit{Metric Learning Approaches}} aim to mitigate domain shift by explicitly minimizing statistical distances between domain distributions. Methods proposed in ~\cite{10979996, 9085896} align second-order statistics in Hilbert space, while other techniques like HoMM~\cite{chen2020homm} perform high-order moment matching. Despite their theoretical grounding, these methods rely on an untenable assumption: that \textit{entire feature vectors should be aligned.} This overlooks the compositional nature of representations, where only specific dimensions encode transferable knowledge. The introduction of domain-specific artifacts poisons the statistical properties of high-dimensional features, distorting their transferability across domains.
\textbf{(3) \textit{Self-supervised Methods}} leverage contrastive learning guided by pretext tasks to acquire domain-invariant representations. Approaches like CLUDA~\cite{ozyurt2023contrastive} progressively extract features via a MoCo architecture enhanced with adversarial training, while other studies~\cite{XU2025125452, 10.5555/3618408.3618926, jin2022domain} employ diverse pretext tasks to promote cross-domain consistency. Critically, \textit{the design of pretext tasks directly governs the quality and transferability of the resulting domain-invariant information.}

Despite their methodological differences, all existing approaches share a common philosophical limitation: they rely on \textit{implicit alignment mechanisms}. Whether via adversarial confusion, statistical distance minimization, or contrastive consistency, these methods presume that global feature alignment will simultaneously achieve domain invariance and preserve semantic content. This lack of explicit control over information preservation constitutes a critical weakness: \textit{adaptation occurs without principled mechanisms to discern what knowledge is retained versus discarded during feature extraction.}

\subsection{E.2. Contrastive Learning}
Contrastive learning has emerged as a powerful representation learning paradigm that maximizes similarity between positive pairs while repulsing negative pairs~\cite{10.1007/978-3-030-58621-8_45, 10.1145/3580305.3599295, 10496248}. 
When applied to domain adaptation, however, it faces a fundamental challenge of defining meaningful cross-domain positive and negative pairs. \textbf{Sampling Bias} and \textbf{False Negatives} constitute a critical limitation in cross-domain contrastive learning. Standard approaches treat samples from different samples as negatives even when they belong to the same semantic class, introducing systematic bias that degrades representation quality~\cite{10.5555/3524938.3525859}. Some methods attempt to address this through debiased sampling~\cite{chuang2020debiased} or supervised contrastive learning~\cite{10.5555/3495724.3497291}. \textbf{Pseudo-labeling Strategies} have been proposed to enable supervised contrastive learning in domain adaptation scenarios~\cite{10130611, litrico_2023_CVPR}. These approaches assign pseudo-labels to target features and use them to define positive/negative pairs. Given that classifier-predicted labels may be biased toward source domain characteristics, potentially causing error accumulation during training, pseudo-label derived from the statistical properties of target features distribution are more reliable.

\end{document}